\documentclass[journal]{IEEEtran}
\usepackage{graphicx}
\usepackage{graphics}
\usepackage{amssymb}
\usepackage{color}
\usepackage{array}
\usepackage{multirow}
\usepackage{amsmath}
\usepackage{bm}
\usepackage{cite}
\usepackage{float}
\usepackage{paralist}
\usepackage[normalem]{ulem}
\usepackage[colorlinks, citecolor=blue]{hyperref}

\newlength\savedwidth
\newcommand\whline{\noalign{\global\savedwidth\arrayrulewidth
                            \global\arrayrulewidth 1pt}%
                   \hline
                   \noalign{\global\arrayrulewidth\savedwidth}}

\ifCLASSINFOpdf
\else
\fi
\begin{document}
%
\title{Stage-Aware Feature Alignment Network for Real-Time Semantic Segmentation of Street Scenes}
%
%
%

\author{Xi~Weng,
        Yan~Yan,~\IEEEmembership{Member,~IEEE,}
        Si~Chen,
        Jing-Hao Xue,~\IEEEmembership{Senior Member,~IEEE,}\\
        and~Hanzi~Wang,~\IEEEmembership{Senior Member,~IEEE}
\thanks{This work was in part supported by the National Natural Science Foundation of China under Grants 62071404 and 61872307, by the Natural Science Foundation of Fujian Province under Grants 2020J01001 and 2021J011185, and by the Youth Innovation Foundation of Xiamen City under Grant 3502Z20206046. \textit{(Corresponding author: Yan Yan.)}}
\thanks{X.~Weng, Y.~Yan, and H.~Wang are with the Fujian Key Laboratory of Sensing and Computing for Smart City, School of Informatics,
Xiamen University, Xiamen 361005, China
(e-mail: xweng@stu.xmu.edu.cn; yanyan@xmu.edu.cn; hanzi.wang@xmu.edu.cn).}
\thanks{S. Chen is with the School of Computer and Information Engineering, Xiamen University of Technology, Xiamen 361024, China (e-mail: chensi@xmut.edu.cn).}
\thanks{J.-H. Xue is with the Department of Statistical Science, University College London, London WC1E 6BT, UK (e-mail: jinghao.xue@ucl.ac.uk).}
\thanks{Copyright \copyright 2021 IEEE. Personal use of this material is permitted. However, permission to use this material for any other purposes must be obtained from the IEEE by sending an email to pubs-permissions@ieee.org.}}

%
%

\markboth{Journal of \LaTeX\ Class Files}%
{Shell \MakeLowercase{\textit{et al.}}: Bare Demo of IEEEtran.cls for IEEE Journals}
%



\maketitle

\begin{abstract}

Over the past few years, deep convolutional neural network-based methods have made great progress in semantic segmentation of street scenes. Some recent methods align feature maps to alleviate the semantic gap between them and achieve high segmentation accuracy.
However, they usually adopt the feature alignment modules with the same network configuration in the decoder and thus ignore the different roles of stages of the decoder during feature aggregation, leading to a complex decoder structure. Such a manner greatly affects the inference speed.
In this paper, we present a novel Stage-aware Feature Alignment Network (SFANet) based on the encoder-decoder structure for real-time semantic segmentation of street scenes. Specifically, a Stage-aware Feature Alignment module (SFA) is proposed to align and aggregate two adjacent levels of feature maps effectively. In the SFA, by taking into account the unique role of each stage in the decoder, a novel stage-aware Feature Enhancement Block (FEB) is designed to enhance spatial details and contextual information of feature maps from the encoder. In this way, we are able to address the misalignment problem with a very simple and efficient multi-branch decoder structure.
Moreover, an auxiliary training strategy is developed to explicitly alleviate the multi-scale object problem without bringing additional computational costs during the inference phase. Experimental results show that the proposed SFANet exhibits a good balance between accuracy and speed for real-time semantic segmentation of street scenes. In particular, based on ResNet-18, SFANet respectively
obtains 78.1\% and 74.7\% mean of class-wise Intersection-over-Union (mIoU) at inference speeds of 37 FPS and 96 FPS on the challenging Cityscapes and CamVid test datasets by using only a single GTX 1080Ti GPU. 

\end{abstract}

\begin{IEEEkeywords}
Real-time semantic segmentation, street scene understanding, deep learning, lightweight convolutional neural network, feature alignment and aggregation.
\end{IEEEkeywords}

%
\IEEEpeerreviewmaketitle

\section{Introduction}
%
%
%
%

\IEEEPARstart{R}{ecent} years have witnessed an increasing interest in the applications of autonomous driving and intelligent transportation. A problem of key importance in these applications is to provide a comprehensive understanding of traffic situations at the semantic level.
{Semantic image segmentation, which assigns a label from a set of predefined classes to each pixel in an image, is a fundamental technique to characterize the contextual relationship of semantic categories in street scenes \cite{Swiftnet}. It can be used as a pre-processing step to remove uninformative regions \cite{remove-1, remove-2} or combined with 3D scene geometry \cite{3D-scene-1, 3D-scene-2}}.
In general, these tasks require not only high-resolution input images to cover a wide field of view, but also fast inference speed for interaction or response.

\begin{figure}[!t]
\begin{center}
  \includegraphics[width=3.5in,  height=1.4in]{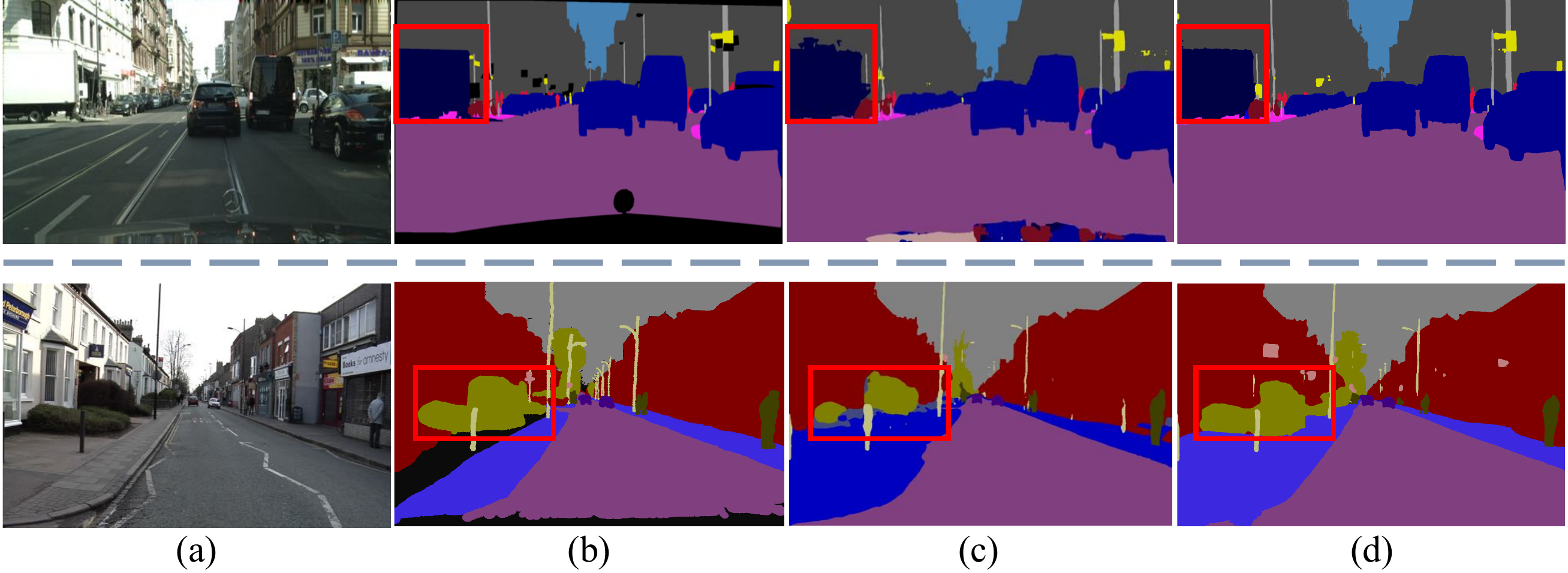}
\end{center}
  \caption{Some segmentation results obtained by SwiftNet \cite{Swiftnet} and our proposed SFANet on Cityscapes \cite{cityscapes} (the first row) and CamVid \cite{camvid} (the second row). The images from left to right denote (a) the input images, (b) the ground-truth images, (c) the outputs obtained by SwiftNet, and (d) the outputs obtained by SFANet, respectively. Compared with SwiftNet, SFANet is able to classify multi-scale objects more accurately and give more refined boundaries (see the red rectangles).}
\label{first-fig}
\end{figure}

{Benefiting from the progress of Deep Convolutional Neural Network (DCNN),  a large number of accuracy-oriented semantic segmentation methods \cite{csvt-Ca-crf, deeplabv3+, pspnet, ocrnet} have been developed and achieved promising performance on a variety of datasets, including street scene datasets (such as Cityscapes \cite{cityscapes} and CamVid \cite{camvid}) and natural scene datasets (such as PASCAL
VOC 2012 \cite{Everingham2010}). These methods exploit rich contextual cues and spatial details to ensure segmentation accuracy.}
However, they often adopt very deep network architectures (such as Xception \cite{xception} and ResNet-101 \cite{resnet}) that involve a large number of network parameters.
Moreover, they take high-resolution images as inputs to capture spatial details, which can result in a heavy memory footprint and large-scale floating-point operations. Therefore, these methods suffer from slow inference speed in practical applications.

Up to date, many efforts have been made towards efficient or real-time semantic segmentation. Some methods either reduce input image resolutions \cite{icnet, erfnet, RTHP} or design highly lightweight network structures \cite{enet, lednet, lrnnet}. Although the above methods greatly reduce the computational complexity of semantic segmentation, they lose the contextual information or spatial details to some extent, thus leading to a significant decrease in accuracy. {Therefore, how to achieve a good balance between inference speed and segmentation accuracy has become a critical challenge in real-time semantic segmentation, especially for complex street scenes.}


Currently, some real-time semantic segmentation methods (such as Bilateral Segmentation Network (BiSeNet) \cite{bisenet} and Deep Feature Aggregation Network (DFANet) \cite{dfanet}) have been proposed. These methods employ a multi-branch framework to combine the contextual information from high-level feature maps with spatial details from low-level feature maps.
In these methods, the high-level feature map is often scaled up to the same size as the low-level feature map, and then an element-wise addition or a channel-wise concatenation operation is used to aggregate two feature maps.
{Unfortunately, the segmentation accuracy of these methods on street scene images, which usually contain many small objects (such as poles and traffic lights), is still far from being satisfactory. This is partly due to the fact that these methods ignore the misalignment between different levels of feature maps, which may lead to the misclassification of boundaries for small objects \cite{alignseg}.}



To mitigate the misalignment problem, Guided Upsampling Network (GUN) is proposed to guide the upsampling of high-level feature maps by using a Guided Upsampling Module (GUM) \cite{gun}. Based on the encoder-decoder structure, Semantic Flow Network (SFNet) \cite{sfnet} develops a Flow Alignment Module (FAM) to take two adjacent levels of feature maps as inputs and learn a semantic flow to align them. However, SFNet adopts the feature alignment modules with the same network configuration at each stage of the decoder. In other words, it ignores that the roles of different stages in the decoder are different (e.g., the early stages of the decoder are concerned with high-level contextual cues, while the later stages of the decoder focus on low-level spatial details). Therefore, SFNet has to rely on a complex decoder structure with multiple shortcut paths to ensure segmentation performance. As a result, its inference speed is influenced.


{Different from natural scene images, street scene images often cover different scales of objects (such as cars and pedestrians). The existence of objects at multiple scales is another important challenge of semantic segmentation that greatly affects the segmentation accuracy in complex street scenes.}
Generally, there are two main strategies to deal with this problem:
1) multi-scale contextual aggregation \cite{deeplabv2, pspnet}, and
2) long-range contextual modeling \cite{ccnet,danet}.
Although these two strategies improve the accuracy, they usually bring additional computational burden for both training and inference.


{In the light of the above issues in street scenes, in this paper, we propose a novel real-time segmentation method, called Stage-aware Feature Alignment Network (SFANet),  through designing an encoder-decoder structure targeting fast inference speed and excellent segmentation accuracy.} A lightweight backbone (ResNet-18) is adopted as the encoder, while a Stage-aware Feature
Alignment module (SFA) is proposed to effectively align and aggregate two adjacent levels of feature maps in the decoder.
In particular, considering the different roles of stages in the decoder, four SFAs with different network configurations (SFA-1, SFA-2, SFA-3, and SFA-4) are developed at all stages of the decoder.

%
%

{In each SFA, we propose a stage-aware Feature Enhancement Block (FEB) to enhance spatial and semantic representations. 
Meanwhile, we adopt a Spatial-Channel Attention module (SCA) to reduce the interference caused by the irrelevant information and preserve the important information.} Based on FEB and SCA, we further use a Feature Alignment and Aggregation module (FAA) to perform feature alignment and aggregation. Hence, the misalignment problem can be greatly addressed.
Moreover, we develop an auxiliary training strategy to explicitly enable each SFA to capture the object information at a certain scale, so as to alleviate the multi-scale object problem in semantic segmentation. It is worth noting that the auxiliary training strategy will not bring additional computational costs during the inference phase since it is only used for training. Some segmentation results obtained by SwiftNet \cite{Swiftnet} and our proposed SFANet are given in Fig.~\ref{first-fig}. Compared with SwiftNet, SFANet is able to segment multi-scale objects more accurately and give more refined boundaries.


In summary, our main contributions in this paper can be summarized as follows:

\begin{itemize}
\item A stage-aware FEB is proposed to enhance the spatial and contextual information of feature maps from the encoder during feature aggregation, while an auxiliary training strategy is introduced to facilitate the training of the network at a certain scale without increasing the computational burden for inference in each SFA. This is beneficial to simultaneously alleviate  the problems of misalignment and multi-scale objects for semantic segmentation of street scenes.

\item Multiple SFAs with different network configurations are elaborately designed to align and aggregate adjacent levels of feature maps. In this way, the spatial details and contextual information can be effectively combined for pixel-level classification, leading to improved segmentation performance of small objects. As a result, a very simple and efficient multi-branch decoder structure can be leveraged to perform real-time semantic segmentation of street scenes.
\end{itemize}

Our proposed SFANet achieves impressive results on two challenging street scene benchmarks. More specifically, based on ResNet-18, our method obtains 78.1\% mIoU and 74.7\% mIoU on the Cityscapes and CamVid test datasets at inference speeds of 37 FPS and 96 FPS, respectively, with a single GTX 1080Ti GPU. This demonstrates that our method is able to strike a good balance between inference speed and segmentation accuracy.

The rest parts of this paper are organized as follows. Section \ref{section:related-work} reviews the related work. Section \ref{section:proposed-method} first gives an overview of the proposed method and then describes the key components in detail. Section \ref{section:experiments} presents and discusses the experimental results. Finally, Section \ref{section:conclusion} draws the conclusion.

\section{Related Work}
\label{section:related-work}
Recently,
a large number of DCNN-based semantic segmentation methods have been proposed and made significant progress. These methods can be roughly divided into two categories: accuracy-oriented semantic segmentation methods and real-time ones. In the following, we briefly review the two categories of methods. Moreover, we introduce some representative feature aggregation methods, which are closely related to our proposed method.

\subsection{Accuracy-Oriented Semantic Segmentation Methods}
Long \textit{et al.} \cite{fcn} develop the pioneering Fully Convolutional Network (FCN), which replaces the fully-connected layers of the classification networks with the convolutional layers, for semantic segmentation. Since then, a large number of accuracy-oriented semantic segmentation methods have been proposed to improve the segmentation accuracy by exploiting spatial details and contextual information of images. 
DeepLab \cite{deeplab} makes use of atrous convolution \cite{atrous-conv} to enlarge the receptive fields without losing spatial resolutions.
U-Net \cite{unet} and DeepLabv3+ \cite{deeplabv3+} adopt the encoder-decoder structure to extract high-level contextual information from the encoder and restore spatial details by the decoder. {Ji \textit{et al.} \cite{csvt-Ca-crf} incorporate cascaded Conditional Random Fields (CRFs) into the decoder to improve the segmentation accuracy. Ji \textit{et al.} \cite{csvt-2-segmentation} propose a deformable DCNN model to capture the semantic information for non-rigid objects.}  Lin \textit{et al.} \cite{refinenet} develop a multi-path refinement network (called RefineNet) to refine feature representations from high-level feature maps to low-level feature maps. {Shi \textit{et al.} \cite{3D-scene-2} leverage a two-stream network to fuse RGB and depth information for RGB-D semantic segmentation. Chen \textit{et al.} \cite{scene-street,scene-street-1} consider different importance levels of distinct classes and design an importance-aware loss for autonomous driving.
Zhou \textit{et al.} \cite{attention-tmm}  introduce two loss functions (i.e., the selection loss and the attention loss) for weakly supervised semantic segmentation.}

One challenge of applying DCNN to semantic segmentation of street scenes is the existence of objects at multiple scales \cite{deeplabv2}. To alleviate this challenge,
DeepLabv2 \cite{deeplabv2} and DeepLabv3 \cite{deeplabv3} use the Atrous Spatial Pyramid Pooling (ASPP)  to capture the multi-scale contextual information. Similarly, Pyramid Scene Parsing Network (PSPNet) \cite{pspnet} develops a Pyramid Pooling Module (PPM) to aggregate feature maps at different pyramid scales.
Recently, a few methods take advantage of non-local operations and self-attention mechanisms to model the contextual relationship between pixels. {Dual Attention Network (DANet) \cite{danet} designs a position and channel attention module to model the dependencies in spatial and channel dimensions.} Criss-Cross Network (CCNet) \cite{ccnet} proposes a novel criss-cross attention module to efficiently encode the contextual information. 

{Although the above methods have achieved impressive performance on various semantic segmentation benchmarks (including street scene ones), they are usually based on deep and wide DCNNs that involve heavy computational operations. This severely hinders the adoption of these methods in the applications demanding real-time inference. Unlike these methods, we adopt a lightweight network to generate different levels of feature maps in the encoder, and multiple SFAs with different stage-aware FEBs to efficiently aggregate the multi-scale spatial and contextual information in the decoder. Such a manner significantly reduces the computational burden for real-time inference.}

\subsection{Real-Time Semantic Segmentation Methods}
During the past few years, real-time semantic segmentation has attracted considerable attention, mainly due to the growing demand for fast inference in many practical applications.
Segmentation Network (SegNet) \cite{segnet} uses a small symmetric structure and the skip connection to accelerate the inference speed. Efficient Neural Network (ENet) \cite{enet} proposes a custom lightweight network and leverages  downsampling operations at the early stages of the network to achieve extremely high inference speed. Similarly, Lightweight Encoder-Decoder Network (LEDNet) \cite{lednet} and Lightweight Reduced Non-Local Operation Network (LRNNet) \cite{lrnnet} design lightweight networks as the decoders, which significantly reduce the number of network parameters. However, these lightweight architectures may affect the learning capability of the network, leading to a performance decrease. Efficient Spatial Pyramid Network (ESPNet) \cite{espnet} and  Efficient Residual Factorized Network (ERFNet) \cite{erfnet} follow the principle of convolution factorization to reduce the computational cost. In particular, ESPNet decomposes the standard convolution into point-wise convolutions and a spatial pyramid of dilated convolutions. ERFNet adopts residual connections and factorized convolutions to balance the trade-off between accuracy and efficiency.

Recently, the multi-branch framework has been developed to achieve real-time semantic segmentation.
Image Cascade Network (ICNet) \cite{icnet} employs a multi-branch architecture to combine cascaded feature maps from different resolution inputs and perform dense prediction. Bilateral Segmentation Network (BiSeNet) \cite{bisenet} adopts a two-branch network to capture the spatial and contextual information, respectively. Note that BiSeNet aggregates the contextual and spatial information at the last stage of the network, which may not fully explore different levels of information in the network, thus affecting the final performance. To address the above problem,
Deep Feature Aggregation Network (DFANet) \cite{dfanet} leverages a feature reuse strategy to achieve fast inference and good accuracy.

In this paper, we also adopt a multi-branch framework to perform real-time semantic segmentation. However, different from existing real-time methods, we elaborately design multiple SFAs with different network configurations to fully exploit feature maps at different stages of the decoder, enabling our network to learn discriminative feature representations for pixel inference. This greatly alleviates the difficulty of extracting rich semantic information due to the adoption of a lightweight network as the encoder.

\begin{figure*}
\begin{center}
\includegraphics[width=0.88\linewidth]{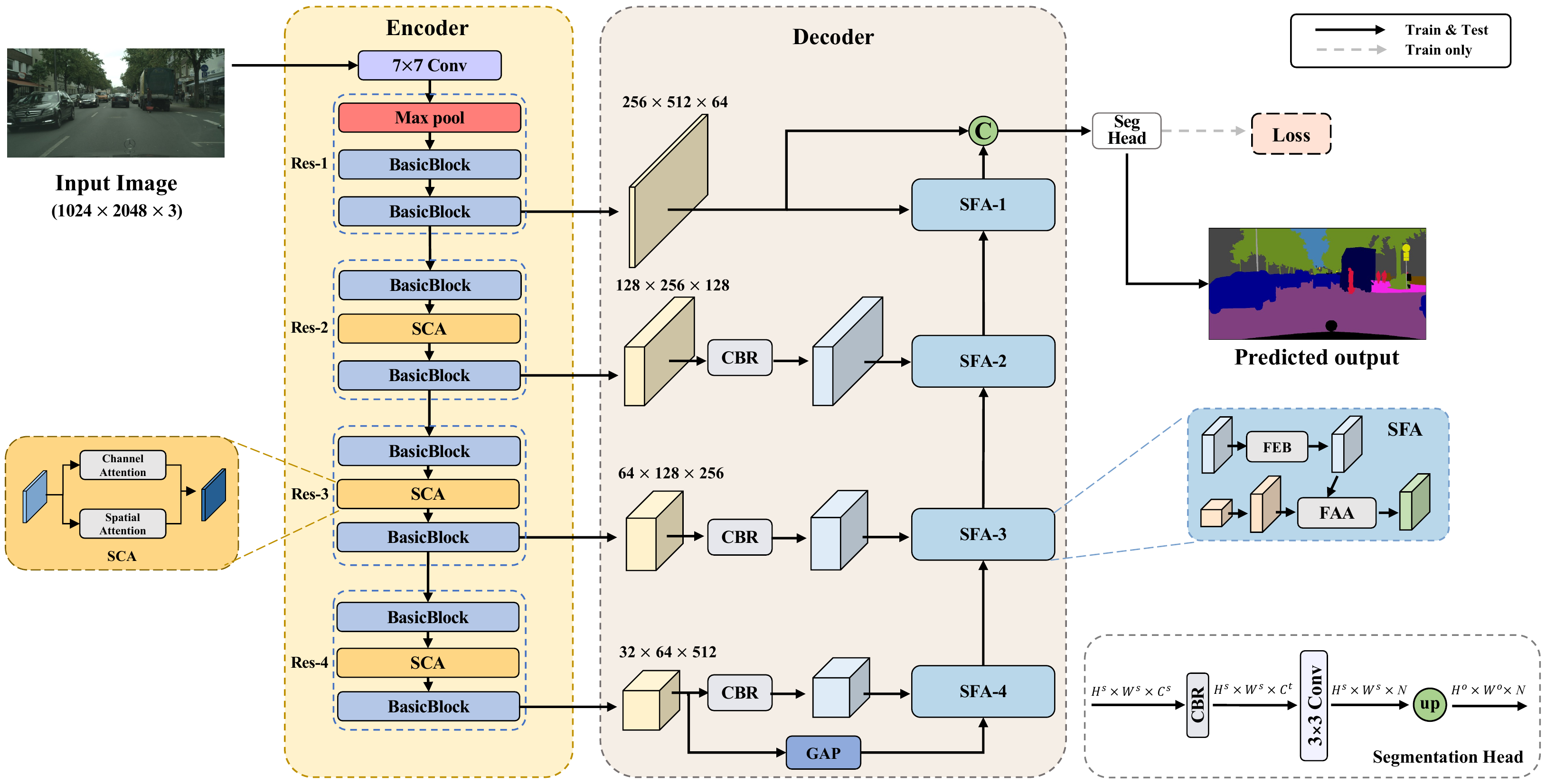}
\end{center}
  \caption{{The network architecture of the proposed SFANet. First, ResNet-18 backbone (incorporated with the Spatial-Channel Attention module (SCA)) with four stages (Res-1 $\sim$ Res-4) offers four different levels of feature maps. Then, the Global Average  Pooling (GAP) operation is employed to capture the global contextual information and the {Conv-BN-ReLU (CBR)} is used to reduce the channel number of feature maps. Next, the feature map obtained from the GAP gradually recovers its spatial details with higher-resolutions by four SFAs (from SFA-4 to SFA-1). Finally, the output feature maps from SFA-1 and Res-1 are concatenated and fed into a segmentation head to generate the final predicted results. In the figure, ``BasicBlock'' represents the basic residual block. ``Seg Head'' means the segmentation head. ``C'' and ``up" denote the channel-wise concatenation operation and the up-sampling operation, respectively.}} 
\label{fig:network-framework}
\end{figure*}

\subsection{Feature Aggregation}

The spatial information is crucial for semantic segmentation. However, the conventional downsampling operations in DCNN may cause the loss of detailed spatial information. Therefore, to achieve satisfactory segmentation performance, many accuracy-oriented and real-time semantic segmentation methods aggregate different levels of feature maps to combine the contextual information with spatial details.

Some methods rely on simple operations  (such as an element-wise addition \cite{bisenet}, \cite{dfanet}, \cite{espnet} or a channel-wise concatenation operation \cite{RTHP}, \cite{Swiftnet}, \cite{edanet}) to perform feature aggregation. However, these methods ignore the misalignment between different levels of feature maps, which may result in poor segmentation performance on the boundaries for small objects.

{To mitigate the above problem, a few recent methods take advantage of feature alignment modules to perform feature aggregation more effectively. }
For example, Guided Upsampling Network \cite{gun} adopts a guided upsampling module to enrich upsampling operators by learning a transformation based on high-resolution inputs.
%
Huang \emph{et al.} \cite{alignseg} propose the Feature-Aligned Segmentation Networks (AlignNet), which mainly consist of an Aligned Feature Aggregation module (AlignFA) and an Aligned Context Modeling  module (AlignCM), to deal with the misalignment problem. Similarly, Semantic Flow Network (SFNet) \cite{sfnet} develops the Flow Alignment Module (FAM) to align and aggregate different levels of feature maps.

In this paper, the proposed SFANet also addresses the misalignment problem between feature maps. However, unlike the above methods, our proposed SFA not only effectively aligns adjacent levels of feature maps, but also enhances the spatial and contextual information during feature aggregation. Such a way enables our method to perform real-time semantic segmentation with a lightweight decoder structure. As a result, SFANet achieves a good balance between inference speed and segmentation accuracy.

\section{Proposed Method}
\label{section:proposed-method}

In this section, we present our proposed SFANet in detail. First, we give an overview of SFANet in Section \ref{Title:Method-architecture}. Then, we describe its critical module SFA in Section \ref{Title:Method-SFA}. Finally, we present some discussions between our proposed method and several related methods in Section \ref{Title:Method-Discussion}.

\begin{figure*}
\begin{center}
\includegraphics[width=0.92\linewidth]{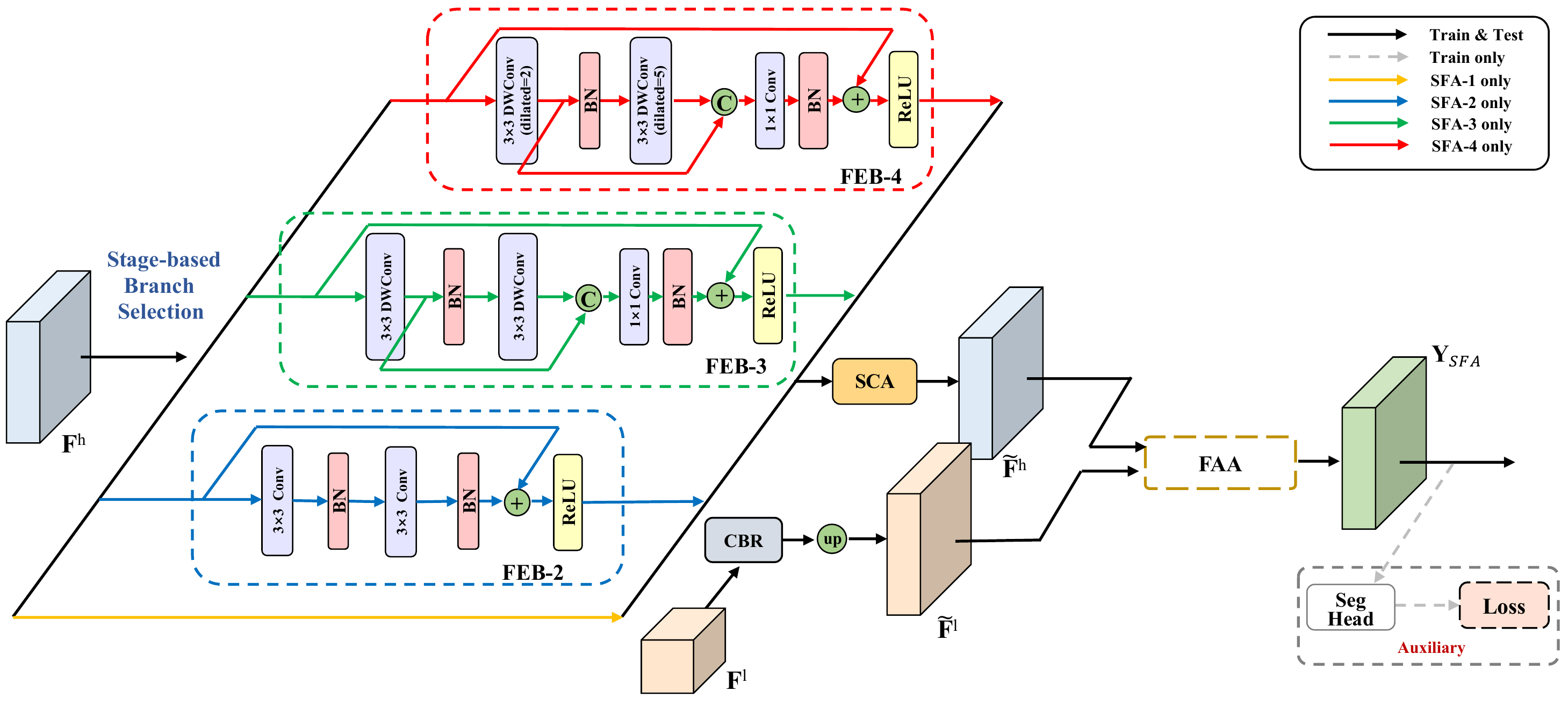}
\end{center}
    \caption{The network architecture of SFA. In the SFA, $\textbf{F}^{h}$ and $\mathbf{F}^{l}$ denote the high-resolution and low-resolution feature maps, respectively. Each FEB with a specific configuration enhances spatial details and contextual information of the high-resolution feature map $\textbf{F}^{h}$ in each SFA. $\widetilde{\mathbf{F}}^{h}$ and $\widetilde{\mathbf{F}}^{l}$ represent the enhanced high-resolution and resized low-resolution feature maps, respectively. $\textbf{Y}_{SFA}$ denotes the output feature map of the SFA. In the Figure, ``DWConv" represents the depthwise separable convolution. ``dilated" represents the atrous rate. ``BN" denotes the batch normalization. ``Auxiliary" indicates the network architecture of the auxiliary training strategy employed in the SFA.} 
\label{fig:SFA_FEB}
\end{figure*}

\subsection{Overview}
\label{Title:Method-architecture}

The network architecture of the proposed SFANet is shown in Fig.~\ref{fig:network-framework}.  SFANet adopts an encoder-decoder structure, where the encoder generates low-level and high-level feature maps while the decoder efficiently combines the contextual information with spatial details to perform pixel-level inference at a low computational cost.

\subsubsection{Encoder}
Due to the requirements of fast inference and small resource consumptions, lightweight DCNNs (such as  ResNet-18 \cite{resnet} and MobileNetV2 \cite{mobilenetv2}) are preferable in real-time semantic segmentation of street scenes. In this paper,
our backbone network (i.e., the encoder) is based on ResNet-18 pretrained on ImageNet \cite{imagenet}.

{More specifically, we remove all the network layers after the last basic residual block of ResNet-18.}
Roughly, ResNet-18 can be divided into four stages (Res-1, Res-2, Res-3, and Res-4), according to the different resolutions of output feature maps \cite{carnet}. In particular, the resolution of an input RGB image is first downsampled to half by using a convolutional layer. Then, the resolution of the downsampled feature map is reduced to one half after passing through each stage. In this way, four different levels of feature maps (corresponding to 1/4, 1/8, 1/16, and 1/32 resolutions of the input image) are extracted from four stages of the backbone network.

Note that the downsampling operations are employed in the first residual blocks of Res-2, Res-3, and Res-4, and thus they may lead to the information loss. Therefore, SCA (see Section \ref{Title:Method-AM}) is incorporated between two consecutive residual blocks of Res-2, Res-3, and Res-4 to improve feature representation capability of the backbone network.

\subsubsection{Decoder}
As shown in Fig.~\ref{fig:network-framework}, the decoder, mainly consisting of four Stage-aware Feature Alignment modules (SFA-1, SFA-2, SFA-3, and SFA-4) and a segmentation head,  effectively and efficiently aggregates feature maps.

More specifically, the channel numbers of the output feature maps from Res-2, Res-3, and Res-4 are first reduced to half by using a Conv-BN-ReLU (CBR) module (i.e., a $3 \times 3$ convolutional layer, followed by a Batch Normalization (BN) layer and a ReLU activation function) to improve the inference speed of the network (in this paper, we do not apply CBR to the output feature map from Res-1 since this feature map can provide rich spatial details for final feature aggregation). Meanwhile, a Global Average Pooling (GAP) operation is performed on the output feature map from Res-4 to capture the global contextual information. Then, four SFAs gradually recover spatial details with higher-resolutions from the input lower-resolution feature maps. Finally, the output feature map from SFA-1 is concatenated with that from Res-1, and the concatenated feature map is fed into a segmentation head to obtain the final predicted results.
In particular, an auxiliary training strategy is proposed in each SFA. This strategy enables the network to explicitly capture objects at different scales and thus further alleviate the problem caused by the existence of objects at multiple scales.

\subsubsection{Segmentation Head}
The details of the segmentation head are illustrated in Fig.~\ref{fig:network-framework}. Suppose that the input feature map is denoted as $\mathbf{I} \in \mathbb{R}^{H^{s} \times W^{s} \times C^{s}}$, where $H^{s}$, $W^{s}$, and $C^{s}$ represent the height, width, and  channel number of the input feature map, respectively.
We first apply a CBR to the input feature map  $\mathbf{I}$ to reduce the channel number from $C^{s}$ to $C^{t}$ to obtain the transition feature map $\mathbf{I}^{t} \in \mathbb{R}^{H^{s} \times W^{s} \times C^{t}}$. Then, the channel number of  $\mathbf{I}^{t}$ is reduced from $C^{t}$ to the final number of semantic classes ($N$) by using a $3 \times 3$ convolutional layer. Finally, an upsamping operation is adopted to generate the feature map $\mathbf{I}^{o} \in \mathbb{R}^{H^{o} \times W^{o} \times N}$, where $H^{o}$ and $W^{o}$ represent the height and width of the final feature map (the same as those of the original input image), respectively. Notice that the channel number $C^{t}$ of $\mathbf{I}^{t}$ can be set to a fixed value to control the computational complexity of the segmentation head. In this paper, we empirically set $C^{t}$ to 64 to achieve a good trade-off between inference speed and segmentation accuracy.


\subsection{{Stage-aware} Feaure Alignment Module (SFA)}
\label{Title:Method-SFA}
To effectively address the misalignment problem during feature aggregation, we design four SFAs with different network configurations (SFA-1, SFA-2, SFA-3, and SFA-4) based on the unique role of each stage in the decoder.
Each SFA aligns two adjacent levels of input feature maps (a high-resolution feature map from the encoder and a low-resolution feature map from its preceding SFA) for feature aggregation.

The network architecture of the proposed SFA is shown in Fig.~\ref{fig:SFA_FEB}. Two feature maps with different resolutions are taken as the inputs of the SFA. The high-resolution feature map $\mathbf{F}^{h} \in \mathbb{R}^{H^{h} \times W^{h} \times C^{h}}$ is from the corresponding stage of the encoder,
and the low-resolution feature map $\mathbf{F}^{l} \in \mathbb{R}^{H^{l} \times W^{l} \times C^{l}}$ is from its preceding SFA.
For the high-resolution feature map in the SFA, it is first fed into a stage-aware FEB and then passed through SCA to obtain the enhanced feature maps, where the important information is emphasized while the irrelevant information is suppressed.
Meanwhile, for the low-resolution feature map, a CBR module and an up-sampling operation are employed to adjust its size to the same as the high-resolution feature map.
In this way, the enhanced high-resolution feature map and the resized low-resolution feature map can be effectively aligned and aggregated by using a Feature Alignment and Aggregation module (FAA). Finally, an auxiliary training strategy is developed to capture the object information at a certain scale.

In the following, we respectively introduce these key components (FEB, FAA, SCA, and the auxiliary training strategy) of SFA in detail.

\begin{figure}[!t]
\begin{center}
  \includegraphics[width=0.7\linewidth]{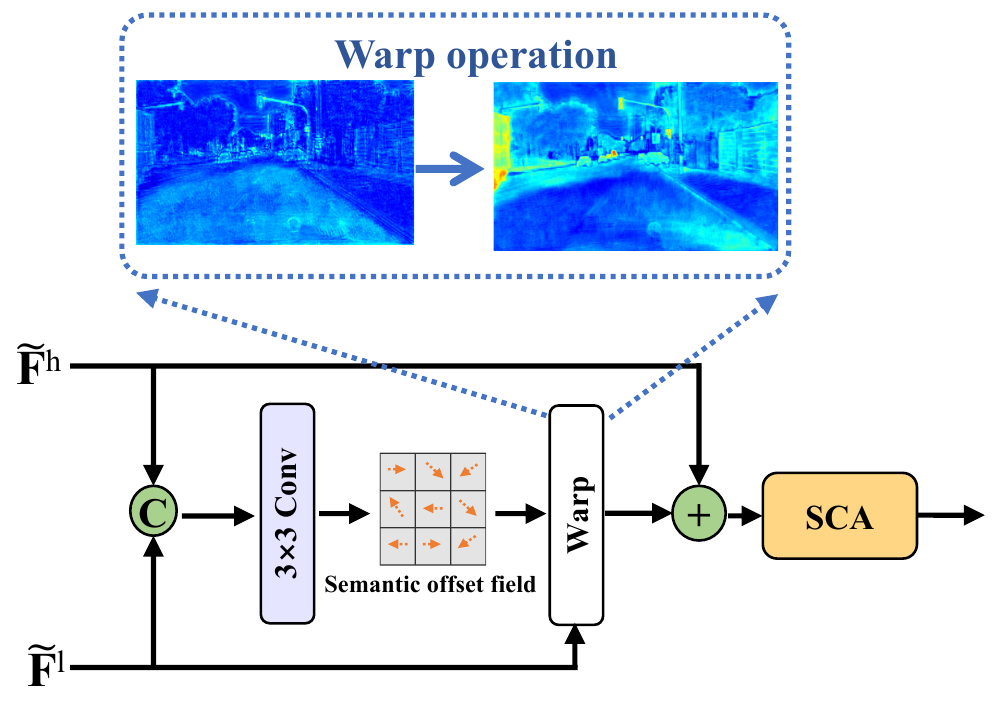}
\end{center}
  \caption{The network architecture of FAA.}
\label{fig:FAA}
\end{figure}

\subsubsection{Stage-Aware Feature Enhancement Block (FEB)}
To ensure real-time inference speed, the lightweight backbone (ResNet-18) is used as our encoder to generate different levels of feature maps. 
However, compared with complex backbones (such as ResNet-101), the model capacity and feature representation capability of ResNet-18 are relatively limited.
Therefore, inspired by residual building blocks \cite{resnet}, a stage-aware FEB is proposed to enhance spatial and contextual representations of feature maps from one stage of the encoder.

Considering the unique role of each stage in the decoder, we design multiple FEBs with different network configurations.
For the early stage (SFA-4), FEB-4 is designed to enhance the contextual information
for the input feature maps from Res-4. For the middle stage (SFA-3), FEB-3 is designed to balance the trade-off between {learning spatial representations} and capturing the contextual information for the input feature maps from Res-3. For the later stage (SFA-2), FEB-2 is designed to enhance {the spatial representations} for the input feature maps from Res-2. Note that we do not use FEB in SFA-1 to avoid a high computational cost because of the large size of  {the} high-resolution feature maps from Res-1.

{The network architecture of FEB is shown in Fig.~\ref{fig:SFA_FEB}.} To be specific, FEB-4 adopts two $3 \times 3$ depthwise separable convolutions \cite{mobilenet}, where the atrous rates of depthwise separable convolutions are respectively set to 2 and 5 to obtain large receptive fields. Mathematically, the above process can be expressed as
\begin{equation}
\begin{split}
\mathbf{F}_{DWC2}&=DWC_{2}(\mathbf{F}^{h}),\\
\mathbf{F}_{DWC5}&=DWC_{5}(BN(\mathbf{F}_{DWC2})), \\
\end{split}
\end{equation}
where $DWC_{2}(\cdot)$ and $DWC_{5}(\cdot)$ represent the depthwise separable convolutions with the atrous rates of 2 and 5, respectively. $BN(\cdot)$ denotes the batch normalization operation.

Then, the two feature maps  (i.e., $\mathbf{F}_{DWC2}$ and $\mathbf{F}_{DWC5}$) are concatenated, which can be formulated as
\begin{equation}
\mathbf{F}_{CAT}=CAT(\mathbf{F}_{DWC2},\mathbf{F}_{DWC5}),
\end{equation}
where $\mathbf{F}_{CAT}$ represents the output feature map. $CAT(\cdot)$ denotes the concatenation operation. By concatenating two different feature maps, $\mathbf{F}_{CAT}$ simultaneously captures two different receptive fields.

Next, we leverage a $1 \times 1$ pointwise convolution to compute channel-wise correlations of concatenated feature maps and batch normalization to output the normalized feature map. Finally, the normalized feature map is combined with the input $\mathbf{F}^{h}$ by a shortcut path and fed into a ReLU layer to obtain the enhanced feature map $\mathbf{F}_{FEB\text{-}4}$. The above process is given as
\begin{equation}
\mathbf{F}_{FEB\text{-}4}=\gamma(\mathbf{F}^{h} + BN(Conv_{11}(\mathbf{F}_{CAT}))),
\end{equation}
where $Conv_{11}(\cdot)$ represents the $1 \times 1$ pointwise convolution. $\gamma(\cdot)$ denotes the ReLU activation function.

FEB-3 uses the same network architecture as FEB-4, except that the atrous rates of depthwise separable convolutions are set to 1 to avoid excessive operational costs and overhead.

FEB-2 employs two standard convolutions to  learn spatial representations for the input feature maps from Res-2. Mathematically, this process can be written as
\begin{equation}
\mathbf{F}_{FEB\text{-}2}= \gamma(\mathbf{F}^{h} + BN(Conv_{33}(BN(Conv_{33}(\mathbf{F}^{h})))),
\end{equation}
where $Conv_{33}(\cdot)$ indicates the standard convolution with the kernel size of $3 \times 3$.

Note that, compared with the basic residual block \cite{resnet}, FEB-2 removes the ReLU layer after the first convolutional layer to preserve more spatial details, which can be beneficial for back-propagating information when training the lightweight network.
FEB-3 and FEB-4 replace the standard convolutional layers in the basic residual block with the dilated convolutional layers. Moreover, they adopt more shortcuts to capture larger receptive fields. Therefore, our stage-aware FEBs can more effectively enhance the spatial and contextual information than the basic residual block for real-time semantic segmentation of street scenes.


\begin{figure}[!t]
\begin{center}
  \includegraphics[width=0.9\linewidth]{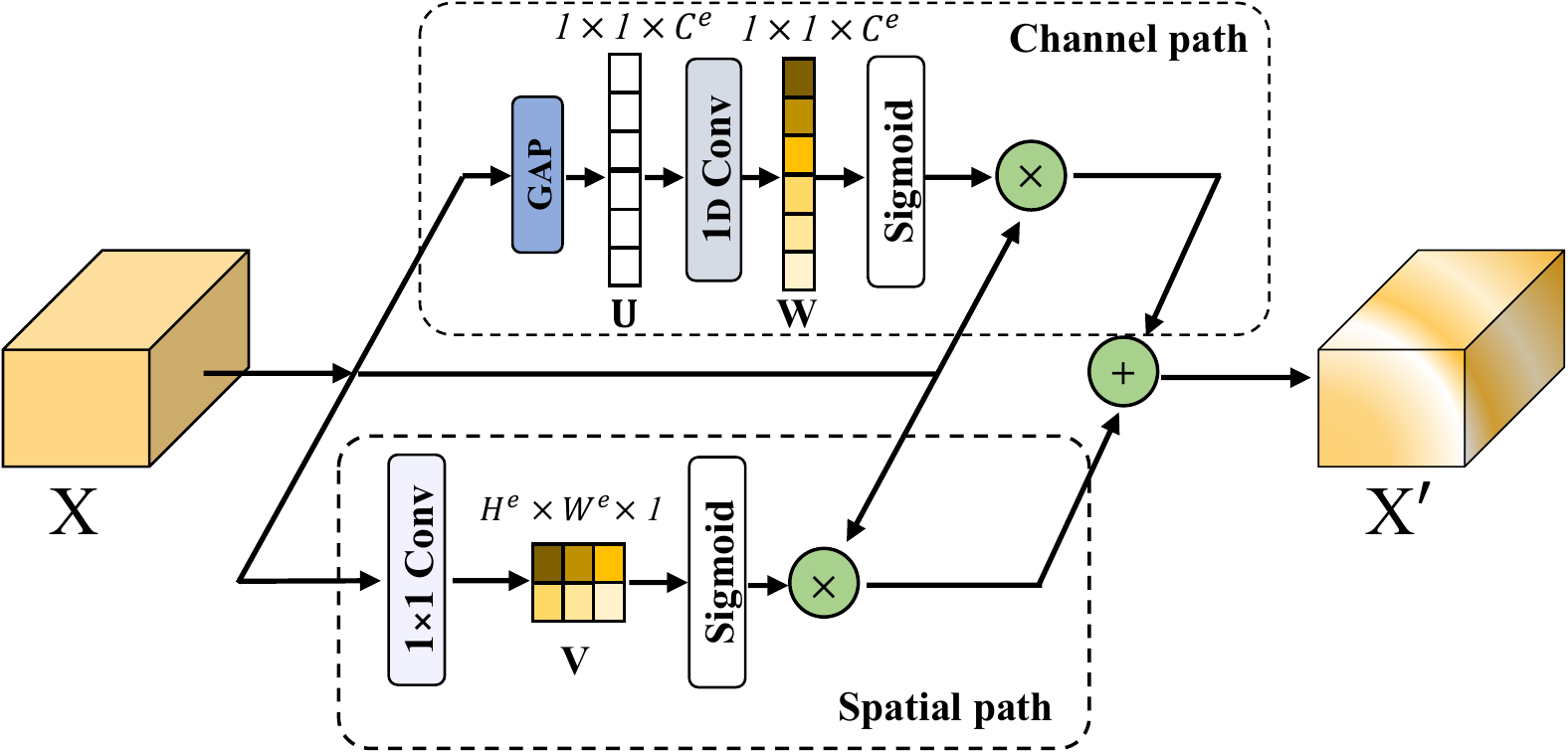}
\end{center}
  \caption{{The network architecture of SCA.}}
\label{fig:AM}
\end{figure}

\begin{figure*}
\begin{center}
\includegraphics[width=0.85\linewidth]{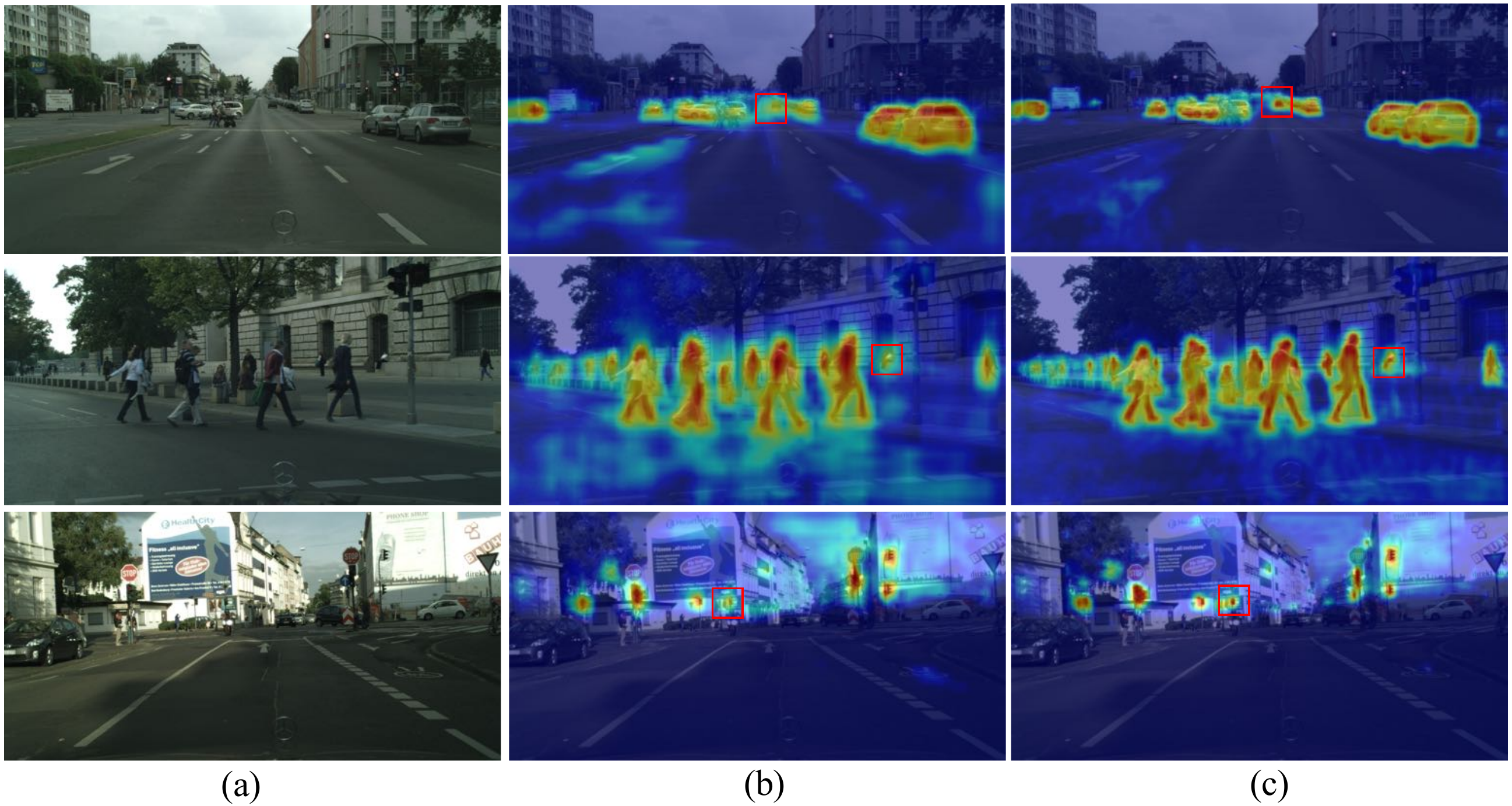}
\end{center}
  \caption{Visualization results of output feature maps from SFA-4 and SFA-3 in the decoder.  (a) The original input images. (b) and (c) show the output feature maps from SFA-4 and SFA-3, respectively. The receptive fields in the later stage of the decoder (e.g., SFA-3) are smaller than those in the early stage of the decoder  (e.g., SFA-4). Thus, SFA-3 focuses on smaller objects than SFA-4 (see the red rectangles).}
\label{fig:Aux_CAM}
\end{figure*}

\subsubsection{Feature Alignment and Aggregation Module (FAA)}
Similar to \cite{alignseg,sfnet}, we use FAA to perform feature alignment and aggregation between the enhanced high-resolution and low-resolution feature maps. The network architecture of FAA is shown in Fig.~\ref{fig:FAA}. Compared with traditional feature alignment modules, we further use SCA to improve the representations of aggregated features.

{Note that FAA is very similar to the optical flow estimation methods \cite{pwc-net, foal}, which compute the motion field from a time-varying sequence of images and are widely used in video-related tasks, including video object detection and motion detection. However, the object position does not change in the semantic image segmentation of street scenes. The differences between two adjacent levels of feature maps are usually subtle in terms of contextual information and spatial details.
In this paper, we develop the stage-aware FEB to enhance the context information and spatial details according to the role of each stage in the decoder. Such a way enables FAA to perform feature alignment and aggregation more effectively, greatly improving the segmentation performance of the network.}
\subsubsection{Spatial-Channel Attention Module (SCA)}
\label{Title:Method-AM}

Recently, attention mechanisms have shown great potential in improving {the} performance of DCNNs, and have been widely used in a variety of computer vision applications {\cite{scSE,senet,ecanet, attention-tmm, cbam}.} In this paper, we also employ a Spatial-Channel Attention module (SCA) to further improve the segmentation performance.

As shown in Fig.~\ref{fig:AM}, SCA is similar to the scSE module \cite{scSE}, except that the channel branch in scSE is simply replaced with the ECA \cite{ecanet} module to effectively remove the disturbance and improve feature representation ability with small resource consumptions.
SCA consists of a spatial path and a channel path. The input feature map $\textbf{X} \in \mathbb{R}^{H^{e} \times W^{e}\times C^{e}}$ is enhanced along the channel and space dimensions to obtain the output feature map $\textbf{X}^{'} \in \mathbb{R}^{H^{e} \times W^{e} \times C^{e}}$.
For the spatial path, SCA first employs a $1 \times 1$ convolution to get a projection tensor \textbf{V} $ \in \mathbb{R}^{H^{e} \times W^{e}\times 1}$. For the channel path, the GAP is employed to obtain the {tensor} \textbf{U} $ \in \mathbb{R}^{1 \times 1 \times C^{e}}$.
Then, a 1D convolution with an adaptive kernel size is used to capture the local cross-channel interaction of \textbf{U}, and thus the tensor \textbf{W} $ \in \mathbb{R}^{1 \times 1 \times C^{e}}$ is obtained as
\begin{equation}
\textbf{W} = \beta \left( \textbf{U} \right),
\end{equation}
where $\beta(\cdot)$ means the 1D convolutional operation.

Finally, a Sigmoid activation function is adopted on both spatial and channel paths to limit the range of the output. The overall attention process can be formulated as
\begin{equation}
\begin{split}
\textbf{X}^{'}_{c} &= \sigma \left( w_{c} \right) \cdot \textbf{X}_{c} + \sigma \left(\textbf{V}\right) \otimes \textbf{X}_{c},
\end{split}
\end{equation}
where `$\otimes$' denotes the element-wise multiplication.
$\textbf{X}_{c}$ and $\textbf{X}^{'}_{c}$ represent the feature maps at the $c$-th channel of \textbf{X} and $\textbf{X}^{'}$, respectively. $w_{c}$ means the weighting factor at the $c$-th channel of $\textbf{W}$. $\sigma \left( \cdot \right)$ denotes the Sigmoid activation function. 

{It is worth pointing out that SCA is different from CBAM \cite{cbam} that also uses spatial and channel attention.
First, the channel attention module in CBAM adopts the design similar to SENet \cite{senet}, where the spatial dimension of the input feature map is compressed to efficiently compute the channel attention. However, such a way may cause the information loss, especially for the lightweight network.  Therefore, we adopt the channel attention module used in ECA \cite{ecanet}, where $1\times1$ convolution is used instead of the squeeze operation, avoiding the information loss.
Second, the spatial attention module in CBAM aggregates spatial information of an input feature map to extract intermediate feature maps by using both average-pooling and max-pooling operations. Then these feature maps are fed into a convolutional layer to generate a spatial attention map. In contrast, in our method,
to achieve good segmentation accuracy while maintaining real-time inference speed, we only use the convolution to generate the final attention map. The channel number in the spatial path of our method is much less than that used in accuracy-oriented segmentation methods. Moreover, without using average-pooling and max-pooling operations, the efficiency of our method can be greatly improved.}

\subsubsection{Auxiliary Training Strategy}

In the decoder, four SFAs with different network configurations effectively aggregate different levels of feature maps, which enables our network to implicitly deal with the multi-scale object problem. {However, such a way still does not fully exploit the capability of the network to capture the multi-scale information.}
Therefore, we further propose an auxiliary training strategy in each SFA to explicitly alleviate the multi-scale object problem. This strategy does not bring the computational burden at the inference phase since it is not used for inference.

Specifically, the feature maps from each SFA are fed into a segmentation head to compute the auxiliary loss, enabling the network to learn the object information at a certain scale.
Mathematically, the auxiliary loss adopts the pixel-wise cross entropy, which is defined as follows:
\begin{equation}
\mathcal{L}_{aux}(\textbf{Z}) = \frac{1}{HW}\sum_{i,j}^{H,W}\sum_{k}^{N}-q_{i,j,k}\log(z_{i,j,k}),
\end{equation}
where $q_{i,j,k}$ = 1 if $k$ is equal to the ground-truth label at the pixel location $(i,j)$, and $q_{i,j,k}$ = 0 otherwise.  $\textbf{Z}$ is the predicted output given by the softmax function. $z_{i,j,k}$ denotes the probability value  at the $(i,j)$-th pixel location and the $k$-th channel of the output. $H$ and $W$ denote the height and width of the predicted output, respectively. $N$ indicates the total number of semantic classes.

Notice that BiSeNetV2 \cite{bisenetv2} adopts a booster training strategy, which employs multiple auxiliary segmentation heads on the semantic branch. However, the proposed auxiliary training strategy and the booster training strategy are intrinsically different. The auxiliary segmentation heads in BiSeNetV2 are used to enhance feature representations on the semantic branch. In contrast, our auxiliary training strategy is used to alleviate the multi-scale object problem. As we mentioned above, the decoder gradually restores the feature maps from the encoder to the final outputs through four SFAs, where the auxiliary training strategy is employed to capture the object information at a scale in each SFA.

\subsubsection{Overall Loss}
The overall loss of our SFANet can be formulated as
\begin{equation}
\mathcal{L} = \mathcal{L}_{p}(\textbf{Y})+\sum_{i=1}^{4}\lambda_{i}\mathcal{L}_{aux}(\textbf{Y}_{SFA\text{-}{i}}),
\end{equation}
where $\mathcal{L}_{p}$ and $\mathcal{L}_{aux}$ denote the principal loss and the auxiliary loss, respectively.  The principal loss also uses the pixel-wise cross entropy same as the auxiliary loss.
 \textbf{Y} denotes the final output of the network. $\textbf{Y}_{SFA\text{-}{i}}$ denotes the output feature map from the SFA-$i$. $\lambda_{i}$ is the balance weight for the auxiliary loss of the SFA-$i$.

Some visualization results are given in Fig.~\ref{fig:Aux_CAM}. As we can see, the output feature maps from SFA-3 are concerned with small objects or boundaries of large objects, while those from SFA-4 focus on large objects. This clearly shows the different roles of SFAs in the decoder.

\subsection{Discussions}
\label{Title:Method-Discussion}
It is worth pointing out that both our proposed method and some previous methods \cite{gun, sfnet, alignseg} leverage feature alignment to improve the segmentation accuracy. However, there are significant differences between our proposed method and these methods.

First, our proposed SFA in SFANet is elaborately designed according to the unique role of each stage in the decoder. As we mentioned before, the early stages of the decoder focus on the contextual information, while the later stages emphasize spatial details. {Therefore, multiple SFAs with different network configurations can better exploit different stages of the decoder.
In each SFA, we develop a stage-aware FEB to enhance the spatial details and contextual information during feature aggregation. By tightly combining several key components (FEB, FAA, and SCA), we are able to align and aggregate different levels of feature maps more effectively.}
As a result, a simple and efficient decoder structure can be used to perform real-time semantic segmentation.
In contrast, previous methods (such as AlignNet \cite{alignseg}, SFNet \cite{sfnet}) use the alignment modules with the same network configuration to perform feature alignment and aggregation at each stage of the decoder. To achieve good segmentation accuracy, these methods either adopt a large DCNN as an encoder \cite{alignseg} to enhance the quality of feature maps, or use dense shortcut connections \cite{sfnet} to promote the information flow. Both ways greatly affect the inference speed.

Second, by taking advantage of multiple stage-aware FEBs and the auxiliary training strategy in four SFAs, our method can simultaneously address the problems of misalignment and multi-scale objects during training. On the contrary, GUN \cite{gun} uses the different sizes of image as the input to alleviate the multi-scale problem, while SFNet \cite{sfnet} adopts additional multi-scale modules (pyramid pooling modules) in the decoder. Such manners often introduce extra computational burden, thereby increasing the inference speed.

\section{Experiments}
\label{section:experiments}
In this section, we perform extensive experiments to evaluate the effectiveness and efficiency of the proposed SFANet.
In Section \ref{Title:EXP-datasets}, we first introduce two representative street scene benchmark datasets and evaluation metrics. Then, in Section \ref{Title:EXP-implementation}, we give the implementation details. Next, in Section \ref{Title:EXP-abalation}, we perform ablation studies to evaluate each component of the proposed SFANet.
Finally, in Section \ref{Title:EXP-comparisons}, we compare the proposed SFANet with several state-of-the-art semantic segmentation methods on the benchmark datasets.

\subsection{Datasets and Evaluation Metrics}
\label{Title:EXP-datasets}
{To show the superiority of the proposed SFANet for semantic segmentation of street scenes, we conduct experiments on the Cityscapes dataset \cite{cityscapes} and the CamVid dataset \cite{camvid}.}

{The Cityscapes dataset is a large-scale urban-scene dataset collected from 50 different cities in Germany. It consists of 25,000 street scene images, where 5,000 images are labeled with fine annotations and 20,000 images are given with coarse annotations.
Each image has the size of 1,024$\times$2,048 and each pixel is annotated to the pre-defined 19 classes. In this paper, we only use 5,000 fine-annotated images for all experiments to validate the performance of our proposed method.
These fine-annotated images can be split into three sets: a training set (containing 2,975 images), a validation set (containing 500 images), and a test set (containing 1,525 images). For a fair comparison, the annotations of the test set are not publicly released for the Cityscapes dataset.}

{The CamVid dataset is another challenging street scene dataset extracted from five video sequences. It contains 701 high-resolution images (each image has the size of 720$\times$960) with high-quality pixel-wise annotations for 11 classes. Following \cite{bisenet, sfnet}, these images can be split into three sets: a training set (including 367 images), a validation set (including 101 images), and a test set (including 233 images).}

{For quantitative evaluation, we adopt the mean of class-wise Intersection-over-Union (mIoU) to measure the segmentation accuracy. For the inference speed, we use Frames Per Second (FPS) to measure the latency. Moreover, we also employ the number of parameters (Params) and float-point operations (FLOPs) to evaluate the memory consumption and computational complexity, respectively.}

\subsection{Implementation Details}
\label{Title:EXP-implementation}
We adopt the PyTorch framework to implement our proposed SFANet.
Instead of training from scratch, we use the publicly available ResNet-18 model pretrained on the ImageNet dataset \cite{imagenet} to initialize our backbone network. All the other weights of our network are randomly initialized by using the Kaiming normal initialization \cite{kaiming-init}.

{To optimize the whole network, we adopt mini-batch Stochastic Gradient Descent (SGD) \cite{sgd} with the momentum of 0.9 and the weight decay of 0.0005 to update the network parameters. The mini-batch size is set to 12 for the Cityscapes dataset and 4 for the CamVid dataset. Moreover, the ``poly" learning rate strategy is employed to decay the initial learning rate, where the initial learning rate is multiplied by $(1-\frac{iter}{total\_iters})^{power}$ at each iteration with the power of 0.9, and it is set to 0.005 and 0.001 for the Cityscapes and CamVid datasets, respectively. As a common practice, we utilize an online hard example mining technique \cite{ohem} to alleviate the influence of class imbalance. We train our model for 120K and 80K iterations for the Cityscapes and CamVid datasets, respectively.}

During the training phase, we employ mean subtraction, random horizontal flip, random scaling (the scale ratio ranges from 0.5 to 2.0), and random cropping to enlarge the training data, as done in \cite{bisenet, dfanet, sfnet}.  During the inference phase, we do not rely on any evaluation tricks (such as sliding-window evaluation and multi-scale testing). Although these tricks can greatly improve the final segmentation results on the test set, they significantly increase the inference speed.

{We use one NVIDIA GTX 1080Ti GPU card to evaluate the inference speed of our method, and repeat 5,000 iterations to mitigate the error fluctuation.
Notice that the BN layers in our network are excluded when we evaluate the inference speed, because they can be merged with previous convolutional layers \cite{Swiftnet}.}

\subsection{Ablation Studies}
\label{Title:EXP-abalation}
{In this subsection, we conduct ablation studies to investigate the effectiveness of each component of our proposed method for real-time semantic segmentation on the Cityscapes validation dataset. All experiments on speed analysis are performed by using the full-resolution image  (with the size of 1,024$\times$2,048) as the input, unless explicitly mentioned.}

\begin{figure}[!t]
\begin{center}
  \includegraphics[width=0.85\linewidth]{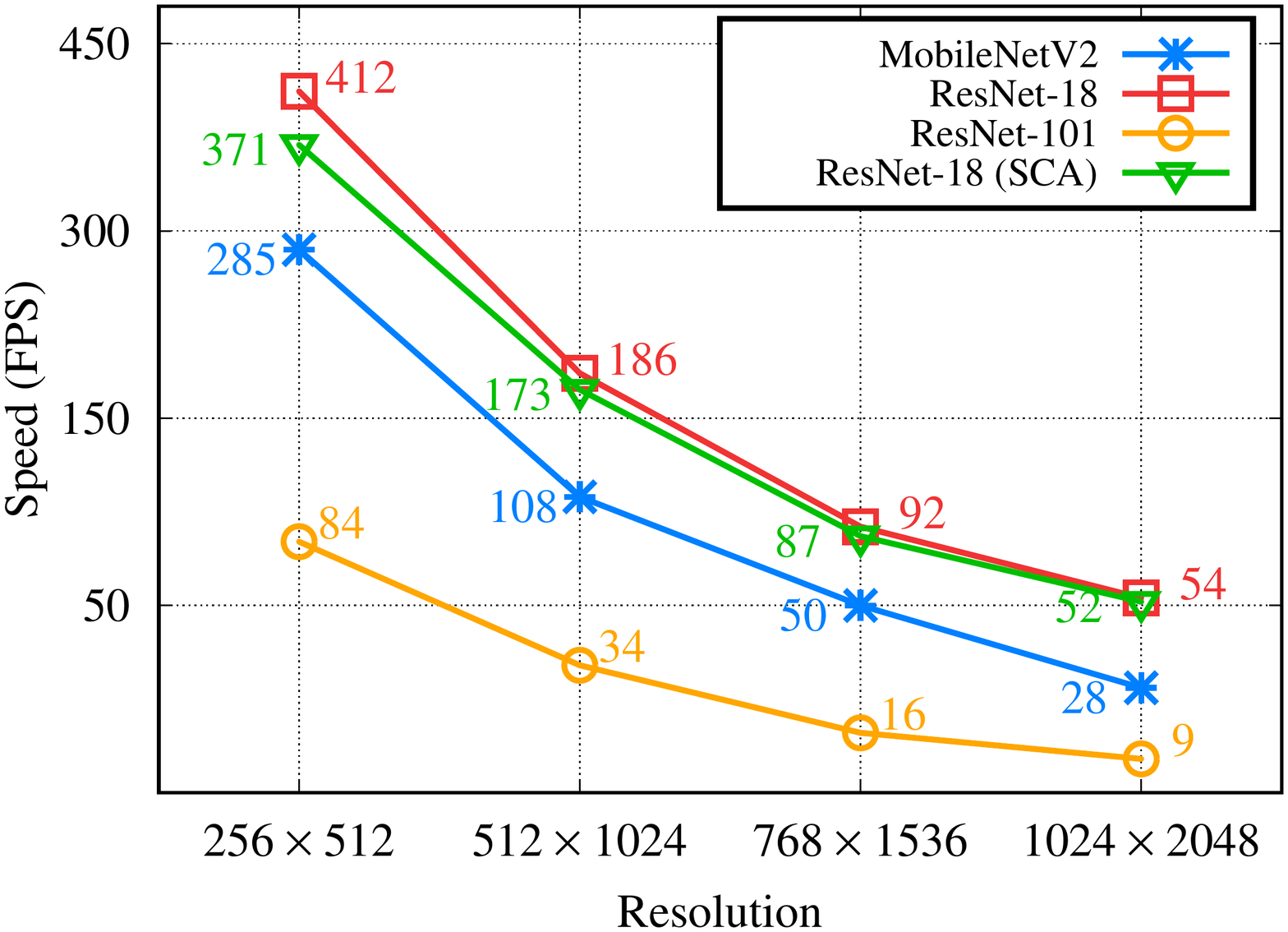}
\end{center}
  \caption{Comparisons of the inference speeds obtained by different backbone networks at different resolution inputs.}
\label{fig:speed-comparison}
\end{figure}

\subsubsection{Ablation for the Backbone Network}
{As we mentioned above, complex DCNNs can provide powerful feature representation capability, but their large numbers of network parameters seriously affect the inference speed. Therefore, lightweight DCNNs are usually adopted as backbone networks for real-time semantic segmentation to ensure efficient inference. In this paper, we use ResNet-18 with SCA, denoted as ResNet-18 (SCA), as our backbone network.}

{To verify the efficiency and effectiveness of  ResNet-18 (SCA), we compare it with three backbone networks, including two lightweight DCNNs (the original ResNet-18 \cite{resnet} and MobileNetV2 \cite{mobilenetv2}) and a complex DCNN (ResNet-101).}

{First, we compare the inference speeds obtained by the backbone networks at different resolution inputs, as shown in Fig.~\ref{fig:speed-comparison}. We can see that ResNet-18 achieves much faster inference speeds than MobileNetV2 at different resolution inputs. ResNet-18 (SCA) obtains the speed similar to ResNet-18. Especially, the inference speeds obtained by ResNet-18 (SCA) and ResNet-18 are very close at the large resolution input. Among all the backbone networks, ResNet-101 achieves the lowest inference speed due to its complex DCNN architecture.}

\begin{table}[!t]
    \caption{Performance Comparisons of Different Backbone Networks: MobileNetV2, ResNet-18, ResNet-101, ResNet-18 (ECA), ResNet-18 (scSE), and ResNet-18 (SCA) on the Cityscapes Validation Dataset.}
    \begin{center}
    \setlength{\tabcolsep}{1.5mm}{
    \renewcommand{\arraystretch}{1.25}
    \begin{tabular}{l|ccc}
        \whline
        Backbone Network & mIoU (\%) & Speed (FPS) & Params (M) \\
        \hline\hline
        FCN+MobileNetV2 & 61.7 & 28 & \textbf{2.04} \\
        FCN+ResNet-18 & 63.6 & \textbf{54} & 11.77 \\
        FCN+ResNet-101 & 65.2 & 9 & 51.95 \\
        \hline
        FCN+ResNet-18 (ECA)  & 63.7 & 53 & 11.77\\
        FCN+ResNet-18 (scSE) & 63.8 & 52 & 12.47\\
        FCN+ResNet-18 (SCA) & \textbf{64.2} & 52 & 11.77\\
        \whline
    \end{tabular}
    }
    \end{center}
    \label{Table:Backbone}
\end{table}

{Next, we further analyze the segmentation performance obtained by different backbone networks on the Cityscapes validation dataset. For a fair comparison, all the backbone networks use FCN-32 \cite{fcn} as the base network structure and are pretrained by ImageNet.
The comparison results are shown in Table \ref{Table:Backbone}.
FCN+MobileNetV2 has the least parameters (about 2.04M Params), but it also achieves the lowest mIoU among all the backbone networks. Meanwhile, FCN+ResNet-101 obtains higher mIoU than the other networks. However, its number of parameters is significantly high (about 51.95M Params).  Although  FCN+ResNet-18 (SCA) gets worse segmentation accuracy (about 1.0\% mIoU lower) than FCN+ResNet-101, it has the much lower number of parameters. Moreover, FCN+ResNet-18 (SCA) achieves about 2.5\% mIoU higher than FCN+MobileNetV2. Compared with FCN+ResNet-18, FCN+ResNet-18 (SCA) attains higher segmentation accuracy. For the speed comparison, FCN+ResNet-101 runs at only 9 FPS in terms of inference speed, mainly due to its high computational cost. FCN+ResNet-18 (SCA) is about 2 times faster than FCN+MobileNetV2 and is close to FCN+ResNet-18 for speed inference.}

\begin{table}[!t]
    \caption{Influence of GAP and SFA on the Performance on the Cityscapes Validation Dataset.}
    \begin{center}
    \setlength{\tabcolsep}{4mm}{
    \renewcommand{\arraystretch}{1.25}
    \begin{tabular}{l|cc}
        \whline
        Method & mIoU (\%) & Speed (FPS) \\
        \hline\hline
        FCN+ResNet-18 (SCA) & 64.2 & 52 \\
        \hline
        Baseline+EA & 74.4 & {45} \\
        Baseline+GAP+EA  & 75.8 & {45} \\
        \hline
        {Baseline+GAP+SFA\_FEB-2} & {77.1} & {38} \\
        {Baseline+GAP+SFA\_FEB-3} & {77.4} & {36} \\
        {Baseline+GAP+SFA\_FEB-4} & {77.8} & {36} \\
        Baseline+GAP+SFA  & \textbf{78.4} & 37 \\
        \whline
    \end{tabular}
    }
    \end{center}
    \label{Table:Decoder}
\end{table}

{Finally, to investigate the influence of different attention modules used in the backbone network, we replace SCA in FCN+ResNet-18 (SCA) with ECA \cite{ecanet} (denoted as FCN+ResNet-18 (ECA)) and scSE \cite{scSE} (denoted as FCN+ResNet-18 (scSE)), respectively, as shown in Table~\ref{Table:Backbone}. These attention modules improve the final segmentation performance. This shows that adopting an attention module in the backbone network can alleviate the disturbance caused by the irrelevant information and highlight the important information.
In particular, FCN+ResNet-18 (SCA) achieves the highest accuracy (about 0.5\% and 0.4\% mIoU higher than FCN+ResNet-18 (ECA) and FCN+ResNet-18 (scSE), respectively) without increasing the number of network parameters. This not only implies the importance of the attention mechanism on the spatial and channel dimensions for semantic segmentation, but also indicates that ECA and scSE can be efficiently combined to achieve better segmentation performance.}

{In summary, the above results show that ResNet-18 (SCA) is able to achieve a good trade-off between speed and accuracy by incorporating SCA into ResNet-18. In the following experiments, we use ResNet-18 (SCA)} as our backbone network (i.e., the encoder in our method).


\subsubsection{Ablation for the Decoder}

{In this subsection, we perform experiments to investigate the influence of key components in the decoder on the accuracy and speed of the proposed SFANet. The performance comparisons are shown in Table \ref{Table:Decoder}. The Baseline+EA method denotes a variant of SFANet, where each SFA is replaced with an Element-wise Addition (EA) operation and the GAP operation is not used.}

{From Table \ref{Table:Decoder}, we can find that the Baseline+EA method achieves the accuracy of 74.4\% mIoU on the Cityscapes validation dataset, and it significantly outperforms FCN+ResNet-18 (SCA) by about 10.2\% mIoU. This verifies the importance of adopting the encoder-decoder structure in real-time semantic segmentation.
By incorporating GAP into the Baseline+EA method, Baseline+GAP+EA obtains better segmentation accuracy  (about 1.4\% mIoU higher) than the Baseline+EA method. By replacing the simple element-wise addition operations with four SFAs, the performance obtained by Baseline+GAP+SFA is boosted by about 2.6\% mIoU and is about 4.0\% mIoU higher than that obtained by Baseline+EA.}

\begin{figure*}[!t]
\begin{center}
\includegraphics[width=0.95\linewidth]{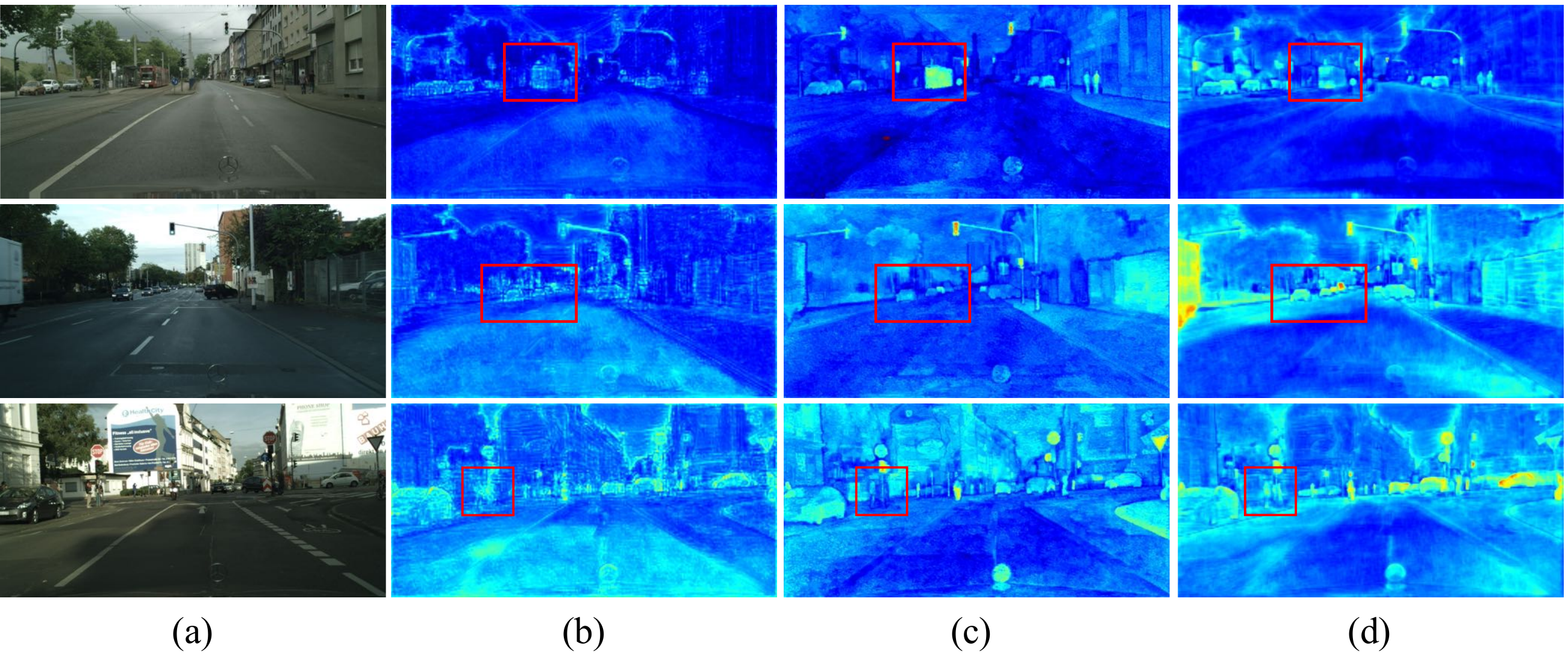}
\end{center}
  \caption{Visualization results of the low-resolution feature map in SFA-1. The images from the first column to the last column denote (a) the input images, (b) the low-resolution feature maps before the warping operation, (c) the low-resolution feature maps after the warping operation (only FAA is used in the SFA), and (d) the low-resolution feature maps after the warping operation (all the key components are used in the SFA).}
\label{fig:SFA_CAM}
\end{figure*}

{The inference speed of the Baseline+EA method is 7 FPS slower than that of FCN+ResNet-18 (SCA). The inference speed of Baseline+GAP+EA runs at 45 FPS. This shows that the GAP can not only effectively exploit the global context information, but also have no much influence on the inference speed. By taking into account both GAP and SFA, the inference speed of Baseline+GAP+SFA is slightly slower than that of Baseline+GAP+EA. 
The above results show that the effectiveness and efficiency of our SFA.}

{In order to further validate the importance of designing different
FEBs at different stages of the decoder, we replace the three stage-aware
FEBs with FEB-2, FEB-3, and FEB-4, respectively. The comparison results are given in Table \ref{Table:Decoder},  where Baseline+GAP+SFA\_FEB-2, Baseline+GAP+SFA\_FEB-3, and Baseline+GAP+SFA\_FEB-4 use the same network structure as Baseline+GAP+SFA, except that the stage-aware FEB in each SFA is replaced with FEB-2, FEB-3, and FEB-4, respectively.}

{From Table \ref{Table:Decoder}, when the same FEB is used in the SFA, the segmentation performance of the network is decreased. Simply using the same module in each stage of the decoder cannot help the network to learn effective discriminative feature maps, thereby affecting the segmentation accuracy. In terms of speed, the speeds of Baseline+GAP+SFA\_FEB-4 and Baseline+GAP+SFA\_FEB-3 are slightly lower than that of Baseline+GAP+SFA, since FEB-4 and FEB-3 have more convolutional layers than FEB-2. Baseline+GAP+SFA\_FEB-2 achieves the fastest inference speed, but its segmentation accuracy is lower than the other two variants. In a word, our SFANet, which designs multiple FEBs at different stages of the decoder, can achieve a good balance between speed and accuracy.}

\subsubsection{Ablation for SFA}
{As mentioned before, our SFA contains four key components (FAA, FEB, SCA, and the auxiliary training strategy). In this subsection, we study the importance of these key components in the SFA, as illustrated in Table \ref{Table:SFA}.} 
We also evaluate the performance of our method when the conventional FAM \cite{sfnet} is used for feature alignment and aggregation instead of FAA.

\begin{table}[!t]
    \caption{Influence of Each Key Component in the SFA on the Final Performance on the Cityscapes Validation Dataset. Here, `Aux' Denotes the Auxiliary Training Strategy.}
    \begin{center}
    \setlength{\tabcolsep}{3.5mm}{
    \renewcommand{\arraystretch}{1.25}
    \begin{tabular}{ccccc|c}
        \whline
        FAM & FAA & FEB & SCA & Aux & mIoU (\%)\\
        \hline
        \checkmark & & & & & 76.2 \\
        & \checkmark & & & & 76.6 \\
        \hline
        & \checkmark & \checkmark & & & 77.6 \\
        & \checkmark& \checkmark & \checkmark & & 77.8 \\
        & \checkmark&\checkmark & \checkmark & \checkmark & \textbf{78.4} \\
        \whline
    \end{tabular}
    }
    \end{center}
    \label{Table:SFA}
\end{table}

From Table \ref{Table:SFA}, by only using FAA in each SFA, our method achieves 76.6\% mIoU and outperforms our method only using FAM  in terms of mIoU (about 0.4\% mIoU higher).
This result shows that  SCA in FAA can increase the final segmentation accuracy due to the enhanced feature representations generated by the attention mechanism.  By combining  FAA and  FEB in SFA, our method achieves the performance of 77.6\% mIoU. This shows that our proposed FEB can effectively enhance spatial and contextual representations of feature maps, leading to performance improvements. By incorporating FAA, FEB, and SCA into SFA, the performance of our method is further improved. Finally, by using all the key components, our method obtains the best segmentation accuracy (about 78.4\% mIoU), demonstrating the effectiveness of the SFA.

We also present some visualization results of the low-resolution feature maps in SFA-1 to show the effectiveness of our proposed SFA, as shown in Fig.~\ref{fig:SFA_CAM}. To be specific, we show the visualization results of low-resolution feature maps before and after the warping operatio. 
Moreover, we also present the visualization results obtained by a variant of SFANet (where only FAA is used in each SFA).
As can be observed, the low-resolution feature maps before the warping operation are blurry and difficult to distinguish different objects.
In contrast, the objects in the low-resolution feature maps after the warping operation are more structured and have more finer boundaries. This shows the importance of feature alignment. Furthermore, by further employing FEB, SCA, and the auxiliary training strategy in the SFA, the low-resolution feature maps are further refined and can focus on the small objects (such as pedestrians and cars) and boundaries. This indicates that FEB, SCA, and the auxiliary training strategy can not only improve the representation ability of the aggregated features in the SFA, but also benefit the feature alignment operation for adjacent levels of feature maps.

\begin{table}[!t]
    \caption{Influence of the Auxiliary Training Strategy in Different SFAs on the Cityscapes Validation Dataset. `\checkmark' Means the Balance Weight of the Auxiliary Loss is Set to 1.}
    \begin{center}
    \setlength{\tabcolsep}{3.8mm}{
    \renewcommand{\arraystretch}{1.1}
    \begin{tabular}{cccc|c}
        \whline
        SFA-1 & SFA-2 & SFA-3 & SFA-4 & mIoU (\%)\\
        \hline\hline
        & & & & 77.8 \\
        \checkmark & & & & 77.3 \\
        & \checkmark & & & 77.8 \\
        &  & \checkmark & & 78.0 \\
        &  &  & \checkmark & 78.1 \\
        \hline
        &  & \checkmark & \checkmark & \textbf{78.4} \\
        & \checkmark  & \checkmark & \checkmark & 78.0 \\
        \checkmark & \checkmark  & \checkmark & \checkmark & 77.4 \\
        \whline
    \end{tabular}
    }
    \end{center}
    \label{Table:aux}
\end{table}

\begin{table*}[!t]
    \caption{Comparisons Between the Proposed Method and Other State-of-the-art Methods on the Cityscapes Test Dataset. `-' Indicates That the Corresponding Result is Not Provided by the Method. FLOPs calculation adopts a 1024 $\times$ 2048 image as the input.}
    \begin{center}
    \setlength{\tabcolsep}{5mm}{
    \renewcommand{\arraystretch}{1.25}
    \begin{tabular}{l|ccccccc}
        \whline
        Method & {Backbone} & Input Size & FLOPs (G) & Params (M) & Speed (FPS) & mIoU (\%)  \\
        \hline\hline
        DeepLab \cite{deeplab} & {ResNet-101} & 512 $\times$ 1024 & 457.8 & 262.1  & 0.25 & 63.1 \\
        PSPNet \cite{pspnet} & {ResNet-101} & 713 $\times$ 713 & 412.2 & 250.8 & 0.78 & 78.4 \\
        \hline
        SegNet \cite{segnet} & {VGG-16} & 640 $\times$ 360 & 286 & 29.5 & 14.6 & 56.1 \\
        ENet \cite{enet} &
        43-layer CNN & 640 $\times$ 360 & 4.4 & 0.4 & 76.9 & 58.3 \\
        ESPNet \cite{espnet} &  9-layer CNN & 512 $\times$ 1024 & 4.7 & 0.4 & 112 & 60.3 \\
        ERFNet \cite{erfnet} &  16-layer CNN & 512 $\times$ 1024 & - & 2.1 & 41.7 & 68.0 \\
        ICNet \cite{icnet} & {PSPNet-50} & 1024 $\times$ 2048 & 29.8 & 26.5 & 30.3 & 69.5 \\
        DABNet \cite{dabnet} & 15-layer CNN & 1024 $\times$ 2048 & - & 0.8 & 27.7 & 70.1 \\
        GUN \cite{gun} & {DRN-D-22} & 512 $\times$ 1024 & - & -  & 33.3 & 70.4 \\
        EDANet \cite{edanet} & 68-layer CNN & 512 $\times$ 1024 & - & 0.68 & 108.7 & 67.3 \\
        LEDNet \cite{lednet} & 55-layer CNN & 512 $\times$ 1024 & - & 0.94 & 71 & 70.6 \\
        DFANet \cite{dfanet} & 43-layer CNN & 1024 $\times$ 1024 & 3.4 & 7.8 & 100 & 71.3 \\
        DF1-Seg \cite{dfseg} & {DF1} & 1024 $\times$ 2048 & - & 8.55 & 80 & 73.0 \\
        DF2-Seg \cite{dfseg} & {DF2} & 1024 $\times$ 2048 & - & 17.2 & 55 & 74.8 \\
        LRNNet \cite{lrnnet} & 55-layer CNN & 512 $\times$ 1024 & 8.58 & 0.68 & 71 & 72.2 \\
        RTHP \cite{RTHP} & {MobileNetV2} & 448 $\times$ 896 & 49.5 & 6.2 & 51.0 & 73.6 \\
        SwiftNet \cite{Swiftnet} & {ResNet-18} & 1024 $\times$ 2048 & 104 & 11.8 & 39.9 & 75.5 \\
        SwiftNet-ens \cite{Swiftnet} & {ResNet-18} & 1024 $\times$ 2048 & 218.0 & 24.7 & 18.4 & 76.5 \\
        SFNet \cite{sfnet} & {DF2} & 1024 $\times$ 2048 & - & 19 & 44 & 77.8 \\
        SFNet \cite{sfnet} & {ResNet-18} & 1024 $\times$ 2048 & 123.5 & 12.87 & 15.2 & 78.9 \\
        BiSeNet \cite{bisenet} & {ResNet-18} & 768 $\times$ 1536 & 55.3 & 49.0 & 45.7 & 74.7 \\
        BiSeNetV2 \cite{bisenetv2} & 18-layer CNN & 1024 $\times$ 2048 & 118.5 & 47.3 & 47.3 & 75.3 \\
        \hline
        \hline
        SFANet (ours) & {DF2} & 1024 $\times$ 2048 & 90.0 & 20.9 & 49 & 78.0\\
        SFANet (ours) & {ResNet-18} & 1024 $\times$ 2048 & 99.6 & 14.6 & 37 & \textbf{78.1} \\
        \whline
    \end{tabular}
    \label{Table:compare-cityscapes}
    }
    \end{center}
\end{table*}

\begin{table}[!t]
    \caption{The Accuracy and Speed Comparisons of the Proposed Method against Other Methods on the CamVid Test Dataset.}
    \begin{center}
    \setlength{\tabcolsep}{3mm}{
    \renewcommand{\arraystretch}{1.25}
    \begin{tabular}{l|cccc}
        \whline
        Method & {Backbone} & Speed (FPS) & mIoU (\%) \\
        \hline\hline
        SegNet \cite{segnet} & {VGG-16} & 46 & 55.6  \\
        ENet \cite{enet} & 43-layer CNN & 61.2 & 51.3  \\
        ICNet \cite{icnet} & {PSPNet-50} & 27.8 & 67.1 \\
        DFANet \cite{dfanet} & 43-layer CNN & 120 & 64.7 \\
        SwiftNet \cite{Swiftnet} & {ResNet-18} & - & 72.6 \\
        SFNet \cite{sfnet} & {DF2} & 105 & 70.4 \\
        SFNet \cite{sfnet} & {ResNet-18} & 41.6 & 73.8 \\
        BiSeNet \cite{bisenet} & {ResNet-18} & 116.3 & 68.7 \\
        BiSeNetV2 \cite{bisenetv2} & 18-layer CNN & 32.7 & 73.2 \\
        \hline\hline
        SFANet (ours) & DF2 & 113 & 74.4 \\
        SFANet (ours) & {ResNet-18} & 96 & \textbf{74.7} \\
        \whline
    \end{tabular}
    }
    \end{center}
    \label{Table:compare-camvid}
\end{table}

\subsubsection{Ablation for the Auxiliary Training Strategy}
{As mentioned above, employing the auxiliary training strategy in SFAs can effectively enable the network to explicitly capture the multi-scale object information, and thus improve the final segmentation performance.
Considering the complexity of the street scenes, in this ablation study, we empirically set the balance weight for the auxiliary loss in each SFA to 0 or 1 to
indicate that whether the corresponding auxiliary loss is used or not.
All the results are shown in Table \ref{Table:aux}.}

{From Table \ref{Table:aux}, we can observe that
employing the auxiliary training strategy in the early stage of the decoder (SFA-4 or SFA-3) is beneficial to improve the segmentation performance (increased by 0.3\% mIoU and 0.2\% mIoU, respectively), compared with SFANet without using the auxiliary training strategy. However, employing the auxiliary training strategy in the later stage of the decoder (SFA-2 or SFA-1) cannot improve the segmentation accuracy of SFANet. On the one hand, in the early stages of the decoder, the scale information is not effectively captured by only using SFA-3 or SFA-4. Therefore, the scale information can be explicitly learned by using the auxiliary training strategy in SFA-3 or SFA-4, improving the final segmentation accuracy. On the other hand, the introduction of the auxiliary training strategy in SFA-1 or SFA-2 affects the learning ability of the network, resulting in performance decline. This is because the scale information captured by the auxiliary training strategy in SFA-1 or SFA-2 is similar to that captured by the segmentation head {in the last stage of the network}. Such a way may increase the difficulty of network learning, leading to a performance decrease.}


{By employing the auxiliary training strategy in both SFA-4 and SFA-3, our method achieves the highest accuracy (about 78.4\% mIoU).
However, the adoption of the auxiliary training strategy in both SFA-2 and SFA-1
is not helpful to improve the segmentation accuracy. Our method using the auxiliary training strategy in {all the SFAs}
is about 1\% mIoU lower than that using the strategy in both SFA-4 and SFA-3. The above results show that the auxiliary training strategy needs to be carefully chosen in each SFA to ensure the final performance.}
\begin{figure*}[!t]
\begin{center}
\includegraphics[width=0.95\linewidth]{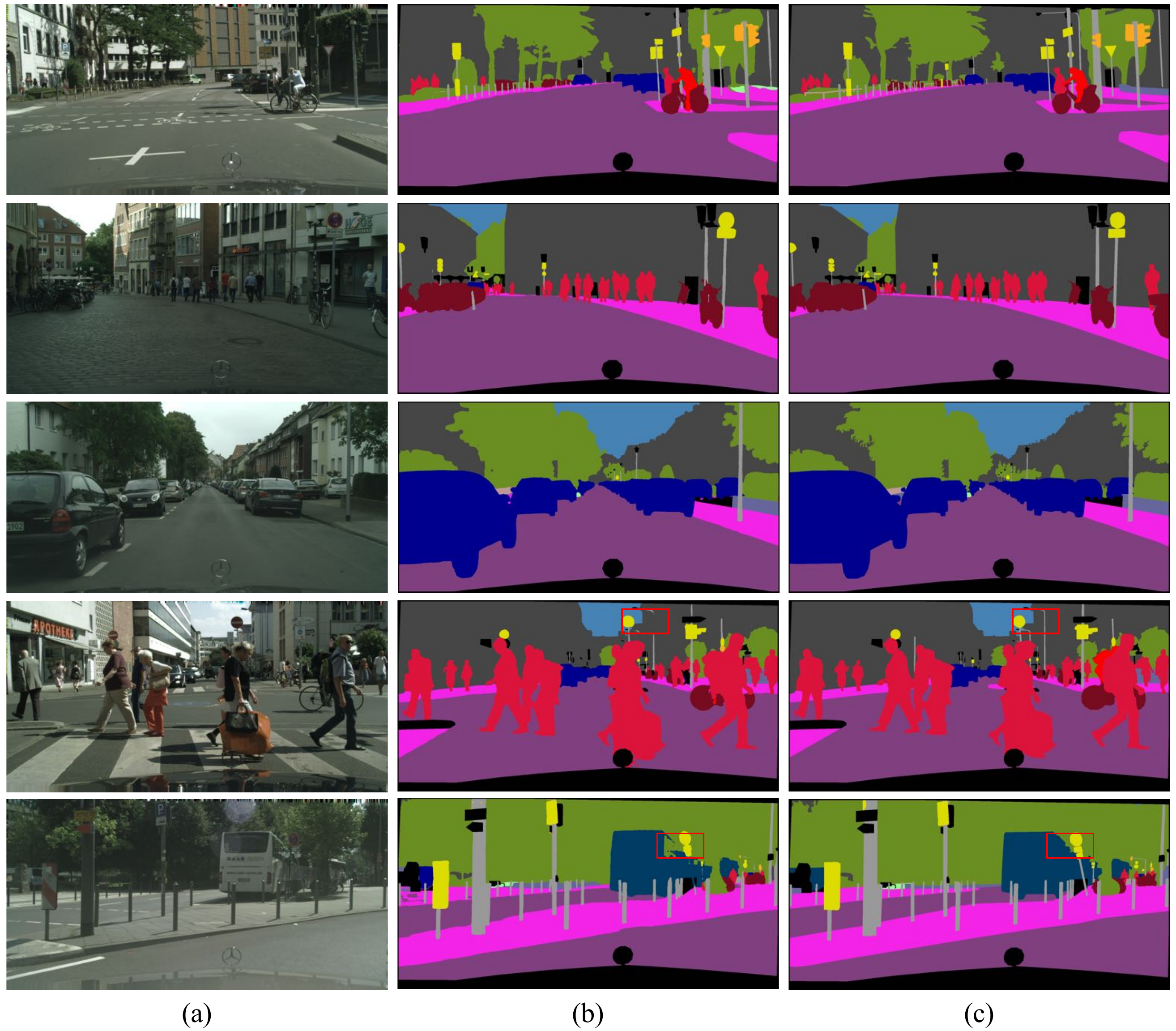}
\end{center}
  \caption{Segmentation results of the proposed SFANet on the Cityscapes validation dataset. The images from the left column to the right column
represent (a) the input images, (b) the predicted results obtained by SFANet, and (c) the ground-truth images, respectively. Some failure cases are shown in the last two rows.}
\label{fig:cityscapes_example}
\end{figure*}

\subsection{Comparisons with State-of-the-Art Methods}
\label{Title:EXP-comparisons}
In this subsection, we compare our proposed SFANet method with several state-of-the-art semantic segmentation methods on the Cityscapes test dataset and the CamVid test dataset, respectively. All the comparison results (including the segmentation accuracy and inference speed) are reported in Tables \ref{Table:compare-cityscapes} and \ref{Table:compare-camvid}. 
All the experiments on measuring the inference speed are not accelerated by TensorRT. Note that we also evaluate the performance of SFANet that adopts DF2 as the encoder to validate the generality of our method. We denote SFANet based on ResNet-18 and DF2 as SFANet (ResNet-18) and SFANet (DF2), respectively.

\subsubsection{Results on Cityscapes}
{Table \ref{Table:compare-cityscapes} shows the comparison results obtained by the proposed SFANet and representative semantic segmentation methods (including several state-of-the-art real-time methods and some accuracy-oriented methods) on the Cityscapes test dataset.}
{Note that state-of-the-art semantic segmentation methods usually leverage different backbones to achieve good segmentation accuracy or fast inference speed, or a good trade-off between them. For fair comparisons between all the competing methods, a number of evaluation metrics (such as mIoU, inference speed, the number of network parameters, and FLOPs) are used to comprehensively evaluate the performance.}

In Table \ref{Table:compare-cityscapes}, we can observe that the proposed SFANet (ResNet-18) and SFANet (DF2) achieve 78.1\% and  78.0\% mIoU at the inference speeds of 37 FPS and 49 FPS, respectively. In particular, SFANet (DF2) gives higher mIoU and faster inference speed than SFNet (DF2). Compared with ESPNet \cite{espnet}, one of the fastest semantic segmentation methods, the proposed SFANet (ResNet-18) is about 17.8\% mIoU higher. Both BiSeNetV2 \cite{bisenetv2} and DFANet \cite{dfanet} also adopt a multi-branch framework for real-time semantic segmentation, but our SFANet (ResNet-18) method obtains better performance. Compared with SwiftNet \cite{Swiftnet} and SFNet (DF2) \cite{sfnet}, our SFANet (ResNet-18) and SFANet (DF2)  not only have less computational complexity and memory consumption, but also achieve higher segmentation accuracy. Furthermore, SFANet (ResNet-18) is even better than an accuracy-oriented method. For example, SFANet (ResNet-18) is about 150 times faster and 15.0\% mIoU higher than DeepLab. Even compared with PSPNet, SFANet (ResNet-18) is only 0.3\% mIoU lower, but 50 times faster.

It is worth pointing out 30 FPS is a common practice to be considered as real-time for semantic segmentation (\cite{gun}, \cite{Swiftnet}, \cite{edanet}). Although the accuracy obtained by SFNet (ResNet-18) is slightly better than that obtained by SFANet (ResNet-18) (about 0.8\% mIoU higher), its inference speed is much lower than 30 FPS and is about two times slower than our method (using the same GPU as ours).
This is because SFNet (ResNet-18) uses a modified ResNet-18, where the partial convolution in the original ResNet-18 is replaced with a convolution layer with a larger kernel size. Although such a way improves the segmentation accuracy, it can affect the inference speed.
Moreover, SFNet (ResNet-18) uses a complex decoder structure.
In contrast, in this paper, we only employ the original ResNet-18 with SCA as the encoder, and design a simple and efficient multi-branch decoder structure. The above results show that our proposed SFANet (ResNet-18) can achieve a good balance between accuracy and speed for real-time semantic segmentation.



{Some qualitative segmentation results are given in Fig.~\ref{fig:cityscapes_example}. We can see that
SFANet is able to correctly assign the semantic labels to the different scales of objects in complex street scenes. However, for some scenarios including severe occlusions and very small objects, SFANet may fail to determine the labels for these objects, leading to wrong segmentation results (see the last two rows in Fig.~\ref{fig:cityscapes_example}). Note that these problems also exist in some state-of-the-art real-time semantic segmentation methods, such as BiSeNetV2 and SFNet.}

\subsubsection{Results on CamVid}
Table \ref{Table:compare-camvid} gives the performance obtained by SFANet and several state-of-the-art semantic segmentation methods on the CamVid test dataset. The proposed SFANet (ResNet-18) and SFANet (DF2) respectively achieve 74.7\% mIoU and 74.4\% mIoU, outperforming the other competing methods. Furthermore, they obtain much faster inference than most of the competing methods. In particular, compared with BiSeNetV2, SFANet (ResNet-18) not only shows  higher mIoU, but also achieves faster inference speed. 
Moreover, the mIoU obtained by SFANet (ResNet-18) is about 0.9\% higher and its inference speed is about 2.7 times faster than SFNet (ResNet-18).

By comparing Table \ref{Table:compare-cityscapes} with Table \ref{Table:compare-camvid}, SFANet (DF2) gives slightly better accuracy (about 0.2\% mIoU higher) and faster speed than SFNet (DF2) for the Cityscapes dataset, while SFANet (DF2) achieves much better performance (about 4.0\% mIoU higher) and faster inference speed than SFNet (DF2) for the CamVid dataset.
Note that our proposed stage-aware FEB is designed to be capable of capturing large receptive fields, thereby facilitating the exploitation of  more spatial and contextual information in the case of small image sizes. Therefore, the mIoU improvement is more evident on the CamVid dataset involving smaller image sizes than on the Cityscapes dataset.
In a word, the proposed SFANet once again demonstrates the excellent performance in terms of the balance between accuracy and inference speed.

Finally, we would like to emphasize that
 our method is not designed only for ResNet-18, since the developed decoder can be combined with any encoder that generates multi-scale feature maps.
 We can see that SFANet (ResNet-18) and SFANet (DF2) achieve good performance on both the Cityscapes and CamVid datasets. Therefore, the developed SFAs are general and can be combined with other backbone networks.


\section{Conclusion}
\label{section:conclusion}
In this paper, we propose a novel SFANet method for real-time semantic segmentation of street scenes. Specifically, an effective SFA is developed to not only alleviate the misalignment problem between two adjacent levels of feature maps, but also explicitly deal with the multi-scale object problem without increasing the computational burden for inference based on an auxiliary training strategy. In particular,
a stage-aware FEB is designed to pertinently enhance spatial and contextual representations of high-resolution feature maps at each stage of the decoder.
Extensive experiments have shown the effectiveness and efficiency of SFANet, which achieves a good balance between accuracy and inference speed.

\ifCLASSOPTIONcaptionsoff
  \newpage
\fi

\begin{IEEEbiography}[{\includegraphics[width=1in]{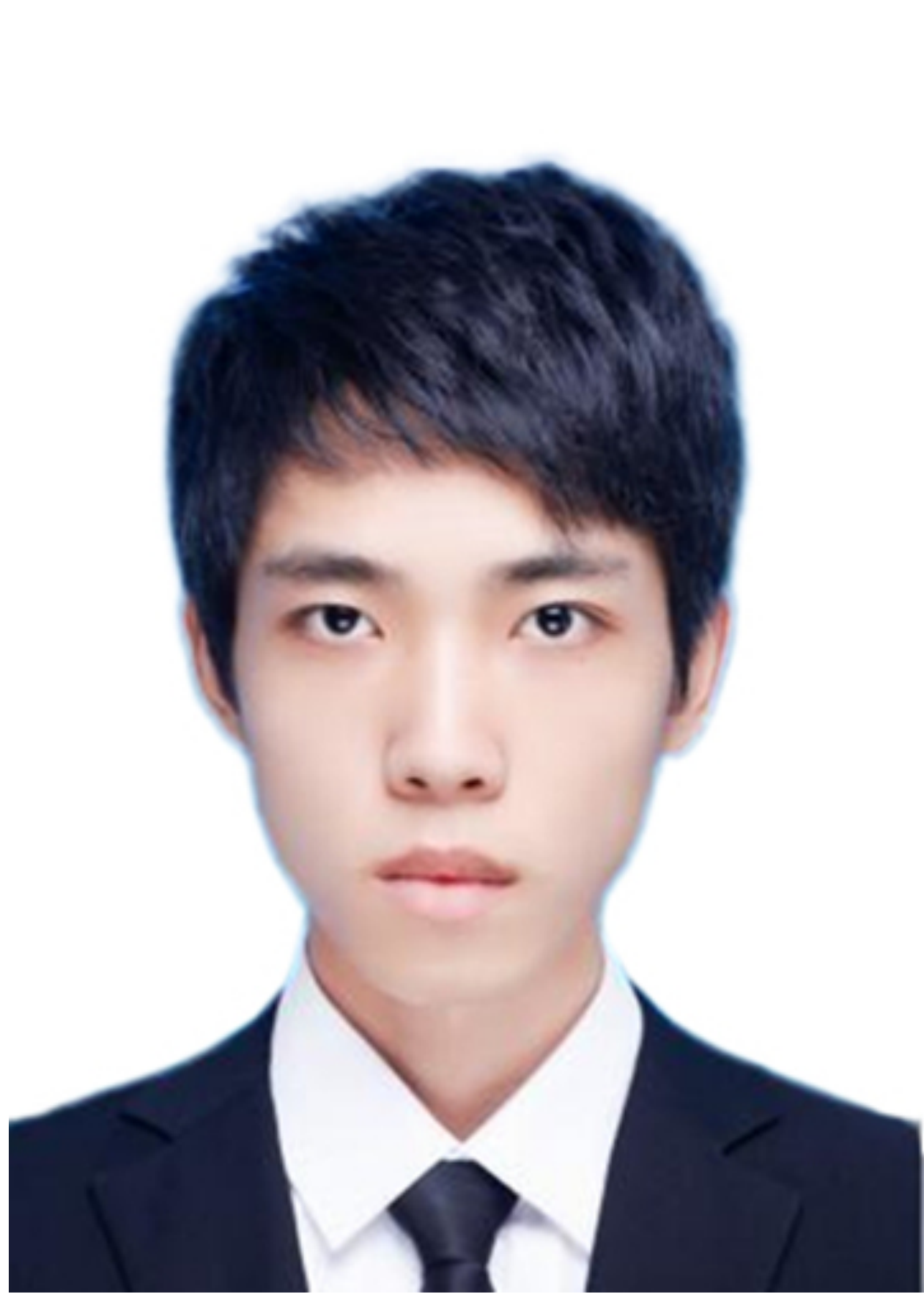}}]{Xi Weng}
is currently a master student in the School of Informatics at Xiamen University, China. His main research interests include deep learning, semantic segmentation, autonomous driving and related fields.
\end{IEEEbiography}

\vspace{-5 mm}

\begin{IEEEbiography}[{\includegraphics[width=1in,height=1.25in,clip,keepaspectratio]{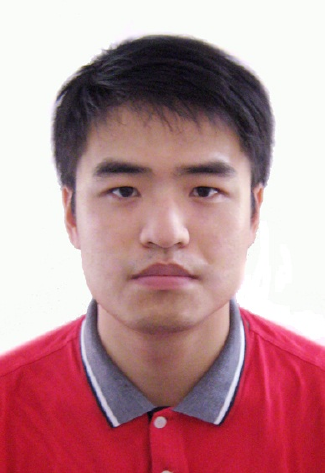}}]{Yan Yan}
is currently a Professor in the School of Informatics at Xiamen University, China.
He received the Ph.D. degree in Information and Communication Engineering from Tsinghua University,
China, in 2009. He worked at Nokia Japan R\&D center as a research engineer (2009-2010) and Panasonic Singapore
Lab as a project leader (2011). He has published around 80 papers in the international journals and
conference including the IEEE T-PAMI, T-IP, T-MM, T-CSVT, T-AFFC, T-CYB, T-ITS, IJCV, CVPR, ICCV, ECCV, ACM MM, AAAI, etc. His
research interests include computer vision and pattern recognition.
\end{IEEEbiography}

\vspace{-5 mm}

\begin{IEEEbiography}[{\includegraphics[width=1in]{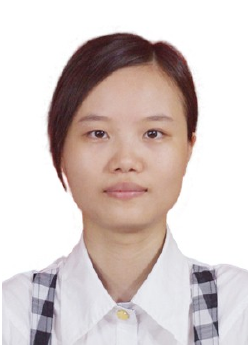}}]{Si Chen} is currently an Associate Professor in the School of Computer and Information Engineering at Xiamen University of Technology, China. She received the Ph.D. degree from Xiamen University, China, in 2014. She won the best student paper award of the 13th China Conference on Machine Learning (CCML) in 2011. Her research interests include computer vision, machine learning and data mining.
\end{IEEEbiography}

\vspace{-5 mm}

\begin{IEEEbiography}[{\includegraphics[width=1in,height=1.25in,clip,keepaspectratio]{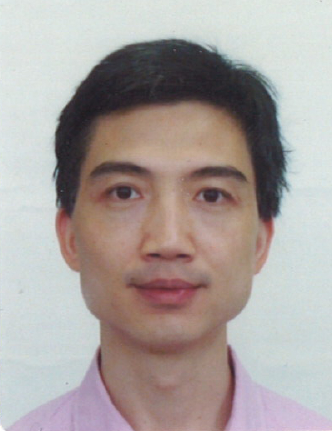}}]{Jing-Hao Xue}
received the Dr.Eng. degree in signal and information processing from Tsinghua University in 1998 and the Ph.D. degree in statistics
from the University of Glasgow in 2008. He is an associate professor in the Department of Statistical Science,
University College London. His research interests include Statistical machine learning, high-dimensional data analysis,
pattern recognition and image analysis.
\end{IEEEbiography}

\vspace{-5 mm}

\begin{IEEEbiography}[{\includegraphics[width=1in,height=1.25in,clip,keepaspectratio]{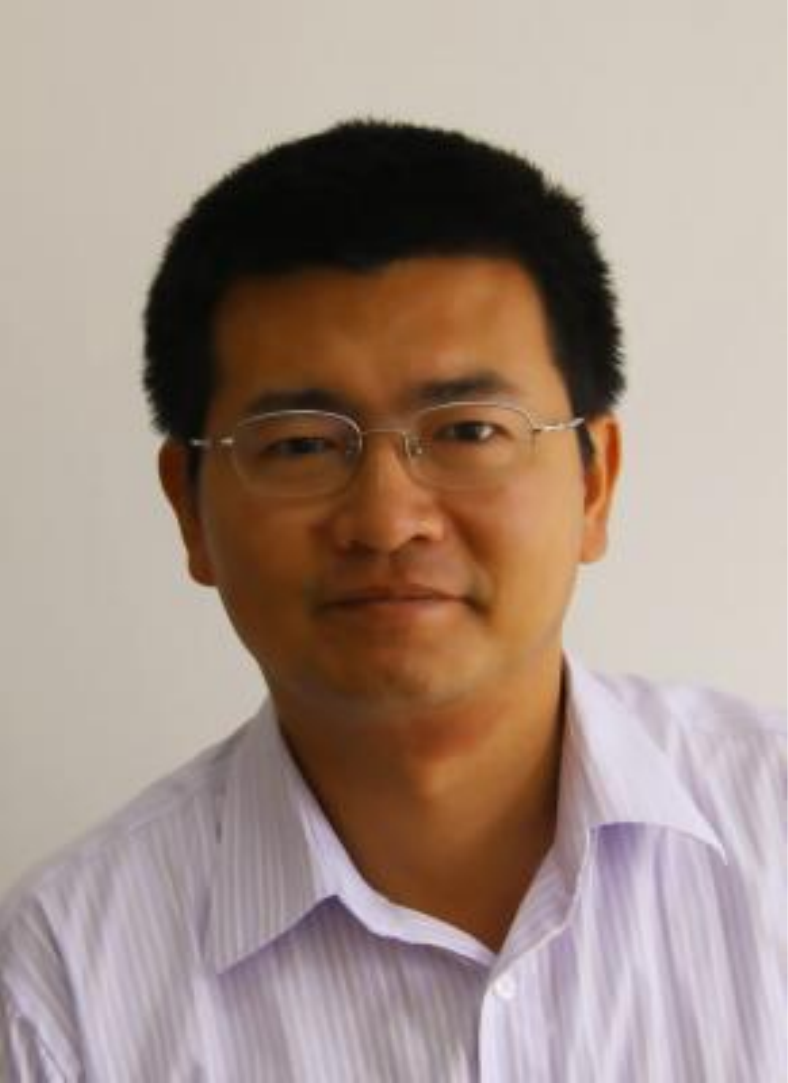}}]{Hanzi Wang}
is currently the head of the Computer Science Department, a Distinguished Professor of "Minjiang Scholars" in Fujian
province and a Founding Director of the Center for Pattern Analysis and Machine Intelligence at Xiamen
University, China. He received his Ph.D. degree in Computer Vision from Monash University, where he was awarded the
Douglas Lampard Electrical Engineering Research Prize and Medal for the best Ph.D. thesis. His research interests include
computer vision and pattern recognition.
\end{IEEEbiography}



\end{document}